\documentclass[letterpaper]{article} 
\usepackage{aaai25}  
\usepackage{times}  
\usepackage{helvet}  
\usepackage{courier}  
\usepackage[hyphens]{url}  
\usepackage{graphicx} 
\usepackage{multirow}
\usepackage{amsmath,amsthm,amsfonts,amssymb,bm,stmaryrd}
\usepackage{longtable}
\usepackage{array}    
\usepackage{color}
\usepackage[table]{xcolor}
\usepackage{arydshln} 
\usepackage{colortbl}
\usepackage{booktabs}
\usepackage{natbib}  
\usepackage{caption} 
\usepackage{algorithm}
\usepackage{algorithmic}

\usepackage{newfloat}
\usepackage{listings}
\usepackage[most]{tcolorbox}
\DeclareCaptionStyle{ruled}{labelfont=normalfont,labelsep=colon,strut=off} 
\floatstyle{ruled}
\newfloat{listing}{tb}{lst}{}
\floatname{listing}{Listing}
\title{Synergistic Multi-Agent Framework with Trajectory Learning for Knowledge-Intensive Tasks}

\author {
    Shengbin Yue\textsuperscript{\rm 1},
    Siyuan Wang\textsuperscript{\rm 2},
    Wei Chen\textsuperscript{\rm 3},
    Xuanjing Huang\textsuperscript{\rm 1}, \\
    Zhongyu Wei\textsuperscript{\rm 1}\thanks{Corresponding author}
}
\affiliations {
    \textsuperscript{\rm 1} Fudan University, Shanghai, China\\
    \textsuperscript{\rm 2} University of Southern California, Los Angeles, USA\\
    \textsuperscript{\rm 3} Huazhong University of Science and Technology, Wuhan, China \\
    sbyue23@m.fudan.edu.cn, sw\_641@usc.edu, lemuria\_chen@hust.edu.cn, 
    \{xjhuang,zywei\}@fudan.edu.cn
}

\usepackage{bibentry}

\begin{document}

\maketitle

\begin{abstract}
Recent advancements in Large Language Models (LLMs) have led to significant breakthroughs in various natural language processing tasks. However, generating factually consistent responses in knowledge-intensive scenarios remains a challenge due to issues such as hallucination, difficulty in acquiring long-tailed knowledge, and limited memory expansion. This paper introduces SMART, a novel multi-agent framework that leverages external knowledge to enhance the interpretability and factual consistency of LLM-generated responses. SMART comprises four specialized agents, each performing a specific sub-trajectory action to navigate complex knowledge-intensive tasks. We propose a multi-agent co-training paradigm, Long Short- Trajectory Learning, which ensures synergistic collaboration among agents while maintaining fine-grained execution by each agent. Extensive experiments on five knowledge-intensive tasks demonstrate SMART's superior performance compared to widely adopted knowledge internalization and knowledge enhancement methods. 
Our framework can extend beyond knowledge-intensive tasks to more complex scenarios. 

\end{abstract}
\begin{links}
\link{Code}{https://github.com/yueshengbin/SMART}
\end{links}

\section{Introduction}
Researchers continue to pursue empowering intelligent systems to generate factually consistent responses in knowledge-intensive tasks~\cite{singhal2022large,yue2023disc,wang2022locate}. Although Large Language Models (LLMs) internalize substantial world knowledge within their parameter memory, they still suffer from fabricating facts, due to their inherent drawbacks, \textit{e.g.}, hallucination \cite{ji2023survey}, trouble in acquiring long-tailed knowledge \cite{kandpal2023large} and struggle to expand their memory \cite{de2021editing}. 
These issues significantly underscore the necessity of incorporating external knowledge from non-parametric (\textit{i.e.}, retrieval-based) memories.

\begin{figure}[t]
    \centering
\includegraphics[width=\linewidth]{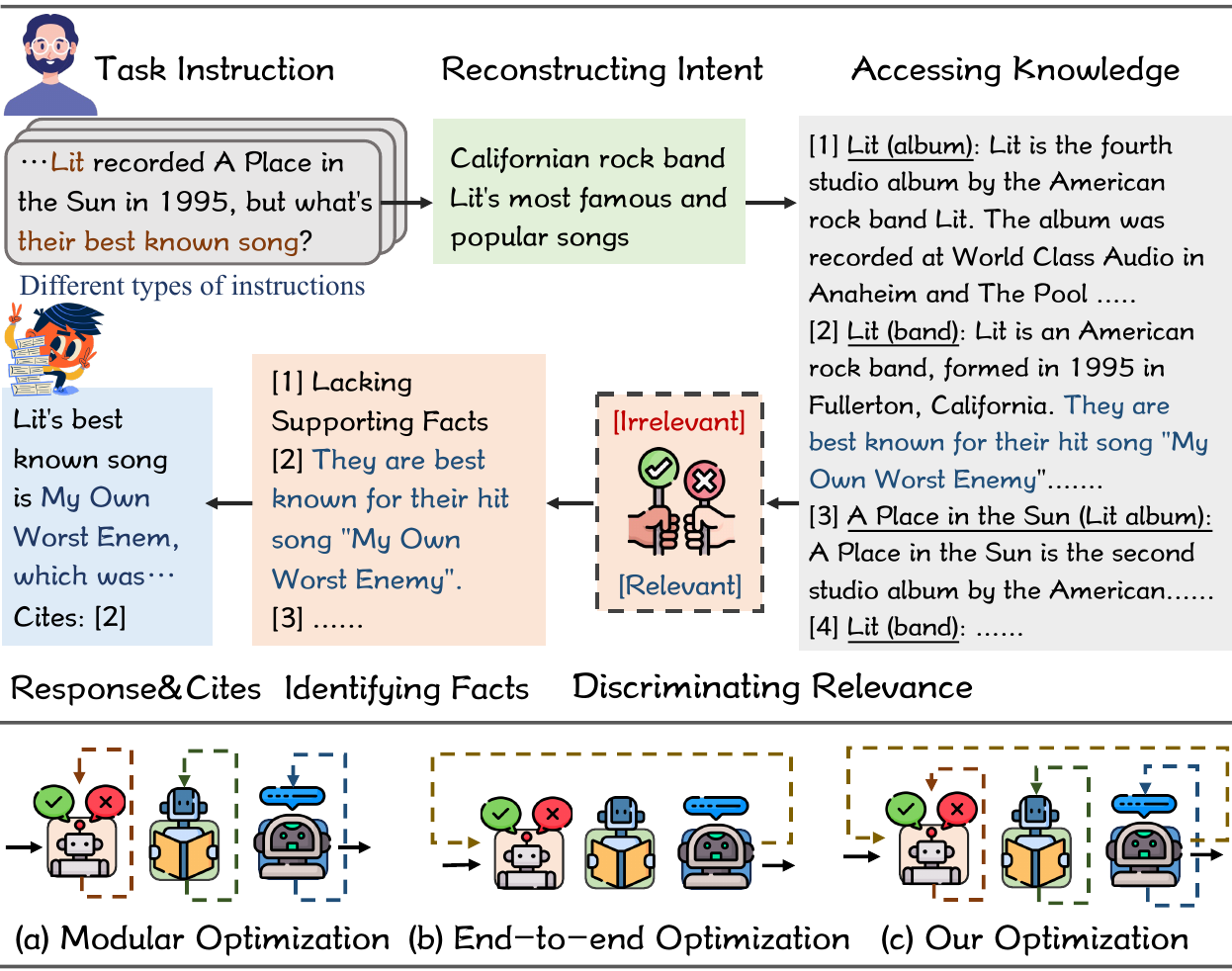}
    \caption{Example of our long trajectory for knowledge-intensive scenarios (Top) and optimization comparison of multi-agent frameworks (Bottom). Solid and dashed arrows indicate inference and optimization paths, respectively.}
    \label{fig:intro}
    \vspace{-3mm}
\end{figure}

Current methods typically augment LLMs with retrieved knowledge to generate responses, which face three main challenges.
(1) \textit{Complex query intent}:
the diverse nature (semantics and form) of instructions 
(\textit{e.g.}, multiple choice, multi-turn dialogue, and complex questions) leads to confusion regarding the query intent of knowledge. 
(2) \textit{Distractors in retrieved knowledge}: knowledge retrieval inevitably introduces noises of varying granularity (document and sentence), with irrelevant documents and superfluous spans distracting the response and resulting in more severe hallucinations.
(3) \textit{Insufficient knowledge utilization}:
LLMs tend to rely more on their implicit knowledge (parameter memory) rather than fully exploiting provided external facts \cite{huang2023survey}. This fact-following disloyalty invalidates the knowledge incorporation process.
Existing knowledge enhancement efforts~\cite{shi2023replug,ma2023query,asai2023self} do not comprehensively address these multi-stage challenges.
To this end, we propose a multi-agent framework, \textbf{S\scalebox{0.8}{MART}}, to integrate different actions to tackle all challenges within complex knowledge-intensive tasks, where each agent performs a specific action.
This comprises an Intent Reconstructor to clarify knowledge intents, a Knowledge Retriever to access external knowledge based on intent, a Fact Locator to evaluate retrieved knowledge and identify factual spans, and a Response Generator that faithfully utilizes and cites available facts.
This process can enhance the knowledge interpretability and response factuality.

However, a major concern remains in how to equip each agent with the necessary capability for corresponding actions while minimizing errors during agent streamline for better overall knowledge-intensive performance. This has been a longstanding challenge in improving multi-agent frameworks, especially as most \cite{yao2023react,hongmetagpt} operate in a non-training manner.
Specifically,
\textit{On one hand}, modular operations, where separate learned modules are pipelined with each dedicated to a specific agent,
can streamline the processing. However,
this can lead to error accumulation as mistakes in earlier modules propagate through the pipeline.
\textit{On the other hand},
encouraging LLM variants to imitate the entire trajectory, while mitigating the fragmentation and error propagation seen in modular systems,
this long-term and global supervision cannot guarantee the precise fine-grained execution by each agent, as it fails to balance the attention each agent devotes to diverse input signals.
Overall, maintaining synergy while ensuring the contribution of various stakeholders is essential.
 
To address this, we propose a multi-agent cooperative training method, namely \textbf{Long Short- Trajectory Learning}, which consists of two stages.
In the first stage, short trajectory learning activates each specific agent in the framework. 
Next, long trajectory learning ensures synergy across multi-agents through trajectory skeleton learning. 
To establish a common supervisory signal for both phases while achieving different training objectives for each, we design special tokens (\textit{i.e.}, trajectory head-end tokens) to allow each agent to identify the attributed trajectories and learn inter-agent interaction signals during training. Specifically, the former phase learns the task output under the prompt of the trajectory-head token, so that the framework learns to distinguish between different agents and confirm the fine-grained information of interest. This independence enables more efficient training with the utilization of existing NLP datasets for pre-training and targeted optimization.
The latter stage requires both predictions of task output and intermittent trajectory tokens throughout the process, \textit{i.e.}, establishing a navigation path from the previous agent to the next.
Our learning approach enables multi-agent systems to collaboratively navigate a long and complex trajectory while concurrently upholding a nuanced representation of each agent.

We conduct experiments on five knowledge-intensive tasks, including fact verification, multiple-choice reasoning, open-domain question answering and long-form generation. Results demonstrate that our framework significantly outperforms pre-trained and instruction-tuned LLMs with more parameters (knowledge internalization methods), and widely adopted knowledge enhancement methods. Further analysis reveals that our long-short trajectory learning enables flexible plug-in combinations of agents while maintaining performance, which is beyond the reach of current end-to-end training systems.
Additionally, the framework achieves impressive performance using only over 40 \% of long trajectory data, substantially reducing the cost and complexity of developing a high-performance multi-agent framework. We envision our framework as a general paradigm that extends beyond knowledge-intensive tasks to more complex scenarios, enabling any multi-agent framework to internalize tailored trajectories.

\begin{figure*}[t]
    \centering
\includegraphics[width=0.99\linewidth]{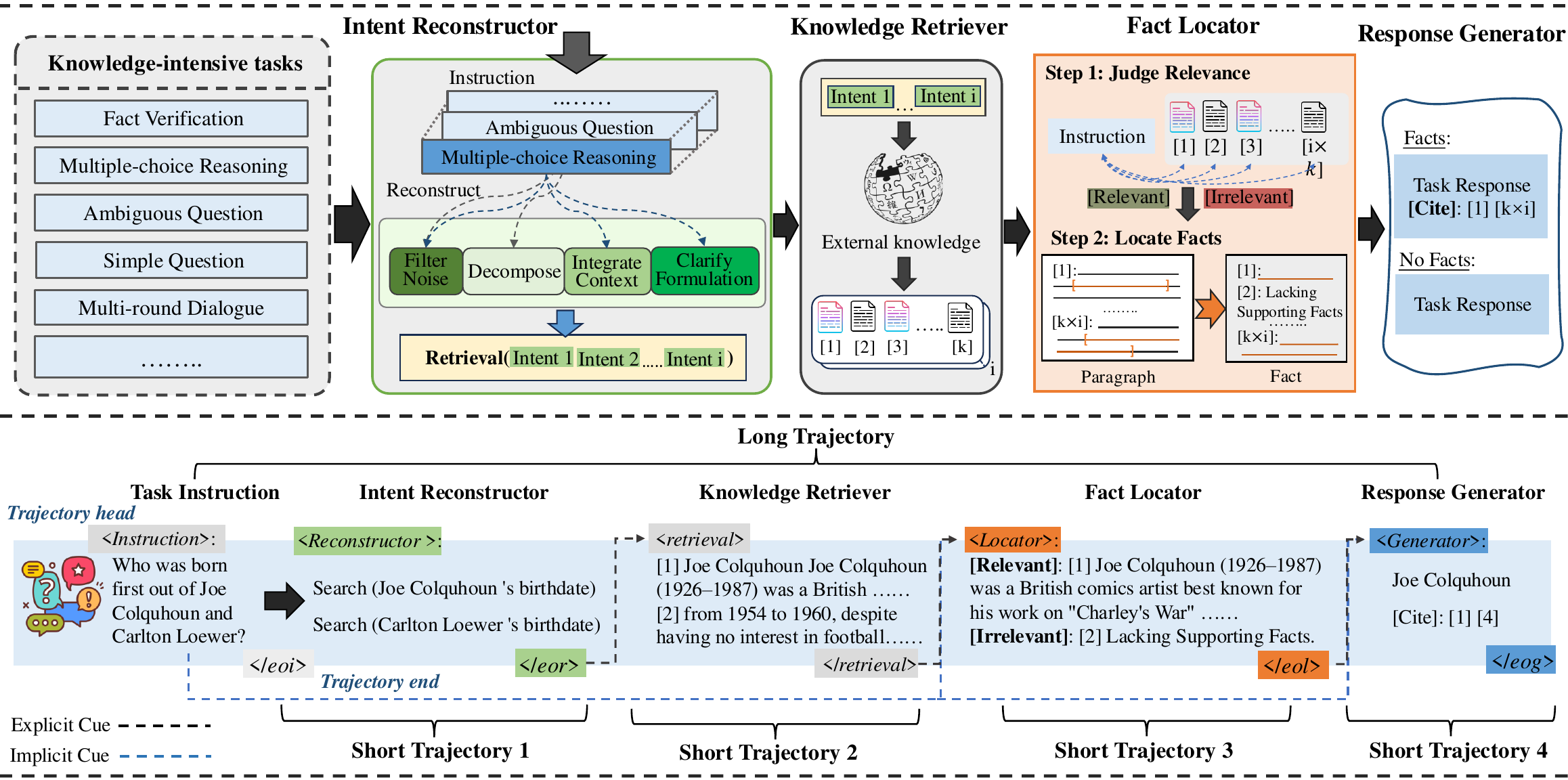}
    \caption{Overview of our multi-agent framework with long- and short-trajectory learning. This framework incorporates four agents: intent reconstructor, knowledge retriever, fac locator, and response generator.}
    \label{fig:model}
\end{figure*}

\section{Method}
\label{method}

Figure \ref{fig:model} provides an overview of our co-framework. We first introduce our multi-agent framework with four key agents performing distinct trajectories. Next, we explain the data construction method and detail the Long-Short Trajectory Learning for optimizing framework synergies.

\subsection{Multi-Agent Framework}
\label{Overview}
To address multi-stage complex challenges in knowledge-intensive scenarios,
we design a multi-agent framework to execute complex long trajectories. This framework incorporates four key agents:
intent reconstructor ($\mathcal{A}_{\mathrm{i} }$), knowledge retriever ($\mathcal{A}_{\mathrm{r} }$), fact locator ($\mathcal{A}_{\mathrm{l} }$), and response generator ($\mathcal{A}_{\mathrm{g} }$).
Each agent serves a specific sub-trajectory, and the final response is obtained by synergizing these agents.

\paragraph{Intent Reconstructor.}
The $\mathcal{A}_{\mathrm{i} }$ agent aims to clarify the knowledge query intent from user instructions. 
It possesses four primary capabilities: integrating contextual clues, identifying key query, unifying task formulation, and intent decomposition, to handle diverse instructions. For example, in multi-turn dialogues, $\mathcal{A}_{\mathrm{i} }$ models long-term history for intent. For noisy instructions, it filters out irrelevant information to identify key queries. For various task formats such as multi-choice QA, $\mathcal{A}_{\mathrm{i} }$ formulate all inputs as a query format for subsequent processing. When handling multi-hop queries like ``Who was born earlier, person A or person B?'', $\mathcal{A}_{\mathrm{i} }$ breaks them down into multiple sub-intents, \textit{i.e.,} each person's birth date. 
By flexibly applying these capabilities, this agent obtains clear query intent to access external knowledge.

\paragraph{Knowledge Retriever.}
The $\mathcal{A}_{\mathrm{r} }$ agent accesses external knowledge bases (e.g., Wikipedia) and obtains relevant knowledge candidates based on reconstructed intents. Specifically,
it is driven by an off-the-shelf retrieval model \cite{izacard2021unsupervised} and acquires top-$k$ knowledge document candidates from the knowledge base for each knowledge intent.
Details of our knowledge retriever setup and the corpus are described in Appendix Sec. B.3.

\paragraph{Fact Locator.}
The $\mathcal{A}_{\mathrm{l} }$ agent aims to locate factual evidence from knowledge candidate sets via document- and sentence-level assessments. Specifically, it assesses the relevance of each knowledge document to the given instruction to determine relevant ones. It then identifies the factual spans from relevant documents as evidence.
The fact locator serves two primary purposes:
1) It enables the agent to check its relevance judgments to minimize the distraction of extraneous spans of the document, and allows the response phase to focus more on fact spans.
2) By explicitly learning to locate facts, it enhances the interpretability of the knowledge application process and bolsters user credibility.

\paragraph{Response Generator.}
The $\mathcal{A}_{\mathrm{g} }$ agent
finally generates responses to user instructions. When facts are provided, it adjusts its knowledge preferences to adhere to them, and ultimately outputs citations to validate loyalty further. In the absence of such information, the response generator relies on its knowledge memory to formulate responses.

\paragraph{Inference Overview.}
The systematic procedure is delineated in the following steps:
$\mathcal{A}_{\mathrm{i} }$  first mines the explicit intent $\bar{q} = \left \{ q_{1},q_{2},...,q_{m}  \right \}$ from the instruction $x$.
Next, $\mathcal{A}_{\mathrm{r} }$ retrieves top-k knowledge documents $\bar{d}=\left \{  d_{1},
d_{2},...,d_{k\times m} \right \} $ using each intent $q_{m}$.
Then, $\mathcal{A}_{\mathrm{l} }$ determines each relevant knowledge passage and further locates the fact span $f \subset d_{k\times m}$. Finally, $\mathcal{A}_{\mathrm{g} }$ utilizes the previous execution trajectory to generate response $y$ and citations when facts exist, otherwise $\mathcal{A}_{\mathrm{g}}$ utilizes only $x$. 
In the $t$-th step, the Agent $\mathcal{A}$ generates a response $r_{t}$ and a head token $h_{t+1}$  of the next trajectory based on the current state of the system:
\begin{equation}
r_{t}, h_{t+1}=\mathcal{A}\left( x,\tau_{t-1}\right),
\label{eq:loss1}
\end{equation}
where $\tau_{t-1} =\left \{ h_{1},r_{1},e_{1},...,h_{t-1},r_{t-1},e_{t-1}\right \}$ denotes the previous execution trajectory. $e$ denotes the trajectory end token. In addition, $\mathcal{A}_{\mathrm{i} }$, $\mathcal{A}_{\mathrm{l} }$ and $\mathcal{A}_{\mathrm{g} }$are built upon same LLMs to fulfill their roles.  The pseudo-code for inference is referenced in Appendix.

\subsection{Trajectory Dataset Construction}
\label{Data}
To implement long-short trajectory learning to optimize our multi-agent framework, we construct the Trajectory dataset.
We collect samples from over 12 knowledge-intensive tasks to ensure coverage of various instruction semantics and formats, such as fact verification~\cite{thorne2018fever}, dialogue~\cite{dinan2018wizard,anantha2021open}, open-domain Q\&A~\cite{kwiatkowski2019natural,stelmakh2022asqa,geva2021did}, and commonsense reasoning~\cite{mihaylov2018can,huang2019cosmos}. Detailed statistics are in Table 5 of Appendix. 
Our dataset contains two components: the long-trajectory subset and the short-trajectory subset.  
The data construction follows two distinct principles:

\textbf{Long-trajectory subset.}
The long-trajectory subset aims to precisely mimic our multi-agent framework inference-time process, which emphasizes the synergy and logical interaction between agents.
Existing work \cite{asai2023self} has demonstrated the effectiveness of the powerful LLM (\textit{e.g.}, GPT3.5, GPT4 \cite{achiam2023gpt}) as a critic model. 
Given an input-output pair $(x, y)$, we create supervised data under the guide of the retrieval ($\mathcal{R}$) and critic model ($\mathcal{C}$). We enable $\mathcal{C}$ to unleash the knowledge intents $\bar{q}$ in $x$ according to the instruction type. 
Then, $\mathcal{R}$ retrieves the top-$k$ knowledge documents based on every $\bar{q}$.
For each document, $\mathcal{C}$ further evaluates whether the passage is relevant based on $(x, y)$. If a passage is relevant, $\mathcal{C}$ further locates and extracts the fact spans. 
Finally, we combine the data and insert the trajectory header and end token 
(\textit{e.g.},\raisebox{-0.2em}{\includegraphics[height=1em]{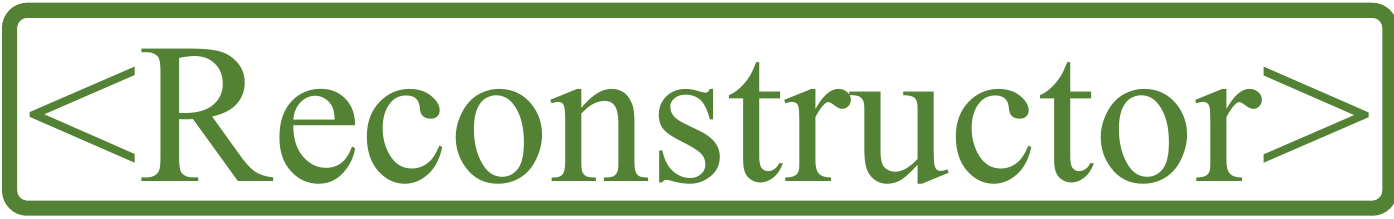}}, \raisebox{-0.2em}{\includegraphics[height=1em]{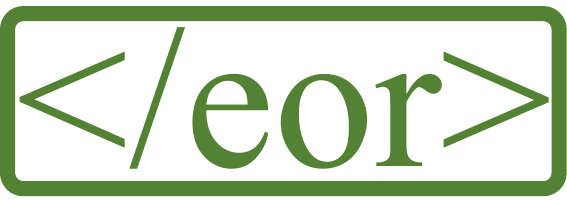}}) into each trajectory.
Trajectory tokens are identifiers that serve as the skeleton of the multi-agent framework. 
In total, we construct 142,507 elaborated instances.

\begin{figure*}[t]
    \centering
\includegraphics[width=0.9\linewidth]{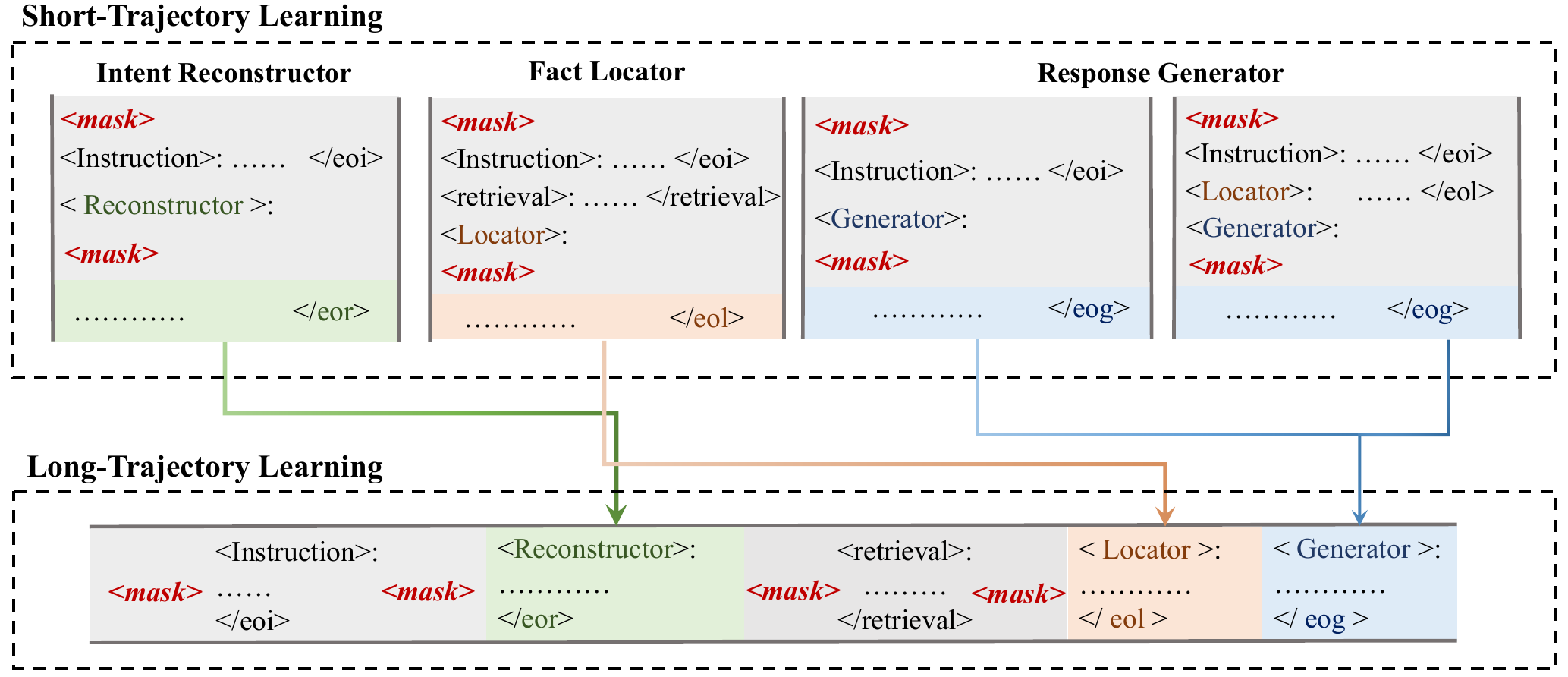}
    \caption{Overview of Long-Short Trajectory Learning. It consists of two stages, for short trajectory learning, under a given trajectory head, requires insight into the various explicit and implicit signals in each particular task. For long-trajectory learning, LLM executes the entire process by predicting different trajectory tokens, ensuring the synergism of different short-trajectories.}
    \label{fig:learnig}
\end{figure*}
\textbf{Short-trajectory subset.}
Unlike the long-trajectory subset, the short-trajectory subset facilitates the training of individual capabilities for each intelligent agent.
This isolation allows us to acquire data directly from a huge amount of existing knowledge-intensive tasks through some simple processing. Thus, we sample from the established NLP and SFT datasets, appending the requisite trajectory header and end token. 
Note that the existing NLP datasets do not fulfill our requirements for intent reconstructing, we employ the methodology utilized in the long-trajectory subset collection. 
Table \ref{tab:token} exhibits the inputs and outputs of each short trajectory under the responsibility of each agent. In addition, the response generator contains two types of inputs to help adapt its knowledge preferences.
We construct a total of 359,791 instances.

\textbf{To summarize.} Two keys are in the construction: the Long-trajectory subset is crafted to emphasize synergy, and the Short-trajectory subset can be easily accessed in large quantities to emphasize uniqueness. 
Refer to Appendix Sec.A for the detail of data construction.

\begin{table}[]
\centering
  \setlength{\tabcolsep}{0.8mm}{
    \scalebox{1}{
\begin{tabular}{lllcc}
\hline
\multirow{2}{*}{Type} & \multicolumn{2}{c}{Trajectory Tokens}                           & \multirow{2}{*}{Input} & \multirow{2}{*}{Output} \\
                                     & Head                               & End                                 &                                 &                                  \\ \hline
$\mathcal{A}_{\mathrm{i} }$                   & \raisebox{-0.1em}{\includegraphics[height=1em]{intent.pdf}}            & \raisebox{-0.1em}{\includegraphics[height=1em]{intent-end.pdf}}      & $x$                              & $\bar{q}$                               \\
$\mathcal{A}_{\mathrm{r} }$                  & \raisebox{-0.1em}{\includegraphics[height=1em]{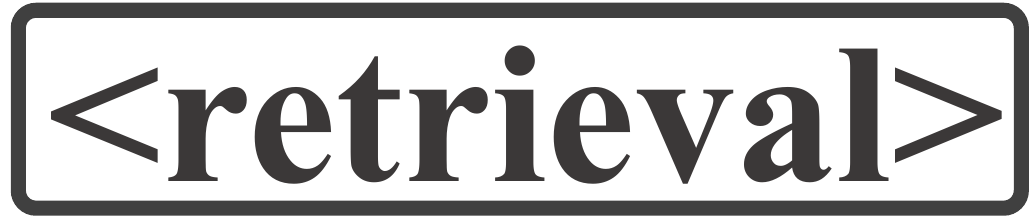}} & \raisebox{-0.1em}{\includegraphics[height=1em]{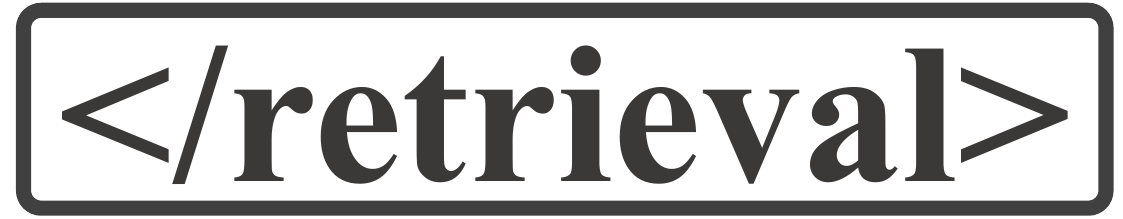}} & $\bar{q}$                               & $\bar{d}$                                \\
$\mathcal{A}_{\mathrm{l} }$                           & \raisebox{-0.1em}{\includegraphics[height=1em]{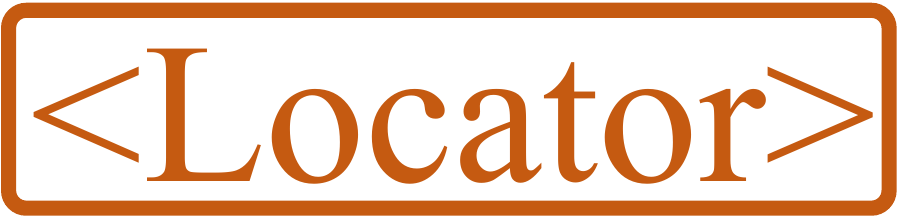}}     & \raisebox{-0.1em}{\includegraphics[height=1em]{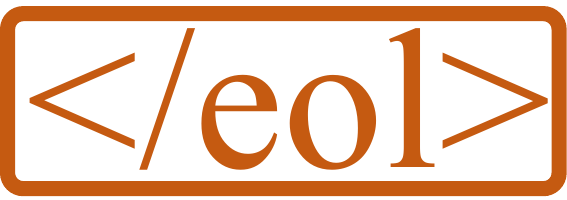}}        & $x$, $\bar{d}$                             & $\gamma$,$\bar{f}$                               \\
$\mathcal{A}_{\mathrm{g} }$                     & \raisebox{-0.1em}{\includegraphics[height=1em]{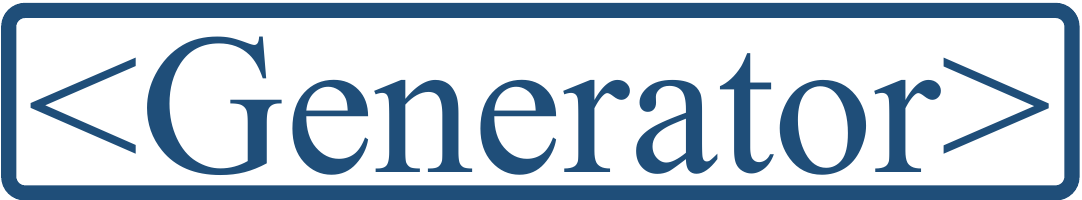}} & \raisebox{-0.1em}{\includegraphics[height=1em]{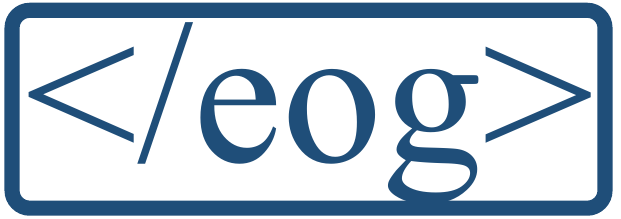}}       & $x$,$\bar{d}$ / $x$                           & $y$                                \\ \hline
\end{tabular}}}
\caption{Four types of trajectory tokens. $x$, $\bar{q}$, $\bar{d}$, $\gamma$, $\bar{f}$ and $\bar{y}$ indicate instruction, intent, knowledge document, relevance tag, fact evidence and response, respectively.}
\label{tab:token}
\end{table}

\subsection{Long-Short Trajectory Learning}
\label{Long Short}
Effectively fine-tuning a trajectory system consisting of multi-agents is a complex task: on the one hand, each agent has its specific trajectory signals of attention. On the other hand, the transformation between different trajectories requires the collaboration of the agents.
In addition, the cost of trajectory data construction for a multi-agent framework greatly hinders the development of such systems.
To this end, we propose Long-Short Trajectory Learning for our multi-agent framework, which consists of two stages, Short Trajectory and Long Trajectory Learning. As shown in Figure \ref{fig:learnig},
Under the guidance of the trajectory head-end token pairs,
the intuition is that Short Trajectory Learning first delineates the responsibilities of each agent to develop their unique capabilities, and then Long Trajectory Learning learns the interactions between them.
This can be understood as initially activating each agent that masters short trajectories within a broader trajectory framework, and then exploring the interconnections between those agents to navigate the full long trajectory.

\begin{table*}[t!]
\centering
\setlength{\tabcolsep}{3.6mm}{
\begin{tabular}{lccccccc}
\hline
\multicolumn{1}{c}{\textbf{Task}}                 & \textbf{Health}           & \textbf{ARC-C}            & \textbf{PopQA}            & \textbf{Squad1} & \multicolumn{3}{c}{\textbf{ASQA}} \\
\multicolumn{1}{c}{Metric}                        & Acc                       & Acc                       & Acc                       & Acc             & Str\_EM     & R-L      & Mauve    \\ \hline
\multicolumn{8}{c}{\cellcolor[HTML]{EFEFEF}\textit{Knowledge internalization
methods}}                                                                                                                             \\
Alpaca2 $_{\mathrm{7B} }\star$    & 44.78                     & 36.43                     & 25.58                     & 11.50           & 14.42       & 28.72    & 51.24    \\
Mistral-Instruct $_{\mathrm{7B} }$              & 65.45                     & 57.84                     & 22.37                     & 14.97           & 20.80       & 32.20    & 33.47    \\
Llama-2-Chat $_{\mathrm{7B} }$                  & 47.95                     & 47.95                     & 25.44                     & 14.13           & 16.79       & 32.35    & 24.21    \\
Vicuna-v1.5 $_{\mathrm{13B} }$                  & 63.01                     & 57.59                     & 17.94                     & 15.25           & 31.95       & 22.99    & 68.41    \\
Llama-2-Chat $_{\mathrm{13B} }$                 & 62.20                     & 48.72                     & 21.22                     & 15.97           & 19.97       & 30.37    & 40.23    \\
ChatGPT                                           & \textbf{76.08}                     & \textbf{77.3}                      & 29.30                     & 22.90            & 39.94       & 35.73    & 44.63    \\ \hline
\multicolumn{8}{c}{\cellcolor[HTML]{EFEFEF}\textit{Knowledge enhancement methods}}                                                                                                                   \\
Alpaca2 $_{\mathrm{7B} }\star$     & 26.44                     & 35.15                     & 33.38                     & 21.41           & 23.59       & 27.21    & 50.09    \\
REPLUG $_{\mathrm{7B} }\star$     & 41.72                     & 47.26                     & 37.24                     & 24.23           & 26.54       & 33.25    & 54.03    \\
VANILLA $_{\mathrm{7B} }\star$    & 29.52                     & 42.74                     & 37.52                     & 25.92           & 32.25       & 34.93    & 39.54    \\
RAIT $_{\mathrm{7B} } \star$      & 52.98                     & 62.10                     & 38.02                     & 23.86           & 25.68       & 15.99    & 12.35    \\
INTERACT $_{\mathrm{7B} }\star$   & 65.45                     & 48.12                     & \underline{ 41.31 }                    &  \textbf{31.52}           & \underline{ 34.54 }      & 35.51    & 43.45    \\
SelfRag $_{\mathrm{7B} }$                        & 68.99                     & 65.52                     & 40.67                     & 22.39           & 28.68       & 34.11    & 83.00    \\ \hdashline
MMAgent $_{\mathrm{3*7B} } \star$ & 70.82 & 63.99 & 36.88 & 23.79                 & 33.04            & \underline{36.49 }         & \underline{88.98}         \\
SMART (OURS)                                       & \underline{ 73.18}                     & \underline{ 65.58}                     & \textbf{42.60}                     & \underline{27.80}           & \textbf{41.16}       & \textbf{40.66}    & \textbf{91.47}    \\ \hline
\end{tabular}
}
\caption{Comparison results against knowledge internalization
and knowledge enhancement methods. $\star$ denotes the method we reproduce based on the same base.
$\star$ denotes re-implemented methods based on the same initial model.
The \textbf{bold} numbers represent the best results and the \underline{ underlined} numbers represent the second.}
\label{tab:main-exp}
\end{table*}
\paragraph{Short Trajectory Learning.}
Short Trajectory Learning is the training of individual capabilities for a single agent. In the context of a long trajectory, it is important to note that short trajectories spanning multiple steps do not necessarily exhibit a strong dependence on preceding short trajectories. To illustrate this point, consider the case of a fact locator, which primarily relies on the original user query and the retrieved results, rather than having a strict dependence on the queries generated in Intent Reconstructor. Similarly, the Response Generator necessitates only the question itself or a combination of the question and the located facts.
As shown in Figure \ref{fig:learnig}, the short trajectory learning first activates each short agent in the framework to focus on the fine-grained signals. 
Given the short-trajectory subset $\mathcal{D}_{\mathrm{short} }=\left \{ \mathcal{D}_{\mathrm{intent}}, \mathcal{D}_{\mathrm{locator} }, \mathcal{D}_{\mathrm{generator} }\right \} 
$,
we initialize a pre-trained LLM  and train it on $\mathcal{D}_{\mathrm{short} }$. 
For each example $\left \{ \left ( x_{i};h_{i} \right )  , \left ( y_{i};e_{i} \right ) \right \} \subset \mathcal{D}_{short}$, 
we use a standard conditional language modeling objective, maximizing likelihood:
\begin{equation}
  \begin{aligned}
\mathcal{L}\left(\mathcal{D}_{short}\right) &= \sum_{i}  \log P_{L M}\left(y_{i};e_{i} \mid  x_{i};h_{i}\right),
\label{eq:loss2}
  \end{aligned}
\end{equation}
Given the inputs and trajectory header, the agent learns to predict the outputs, \textit{i.e.}, delineate different belonging trajectories for the agent to make them understand the fine-grained representations of the corresponding tasks.
This phase utilizes easily accessible and extensive data to build the basic capabilities of the trajectory, reducing the cost of such a framework while maintaining the creativity and versatility of the agent.

\paragraph{Long Trajectory Learning.} 
After the above stage, the framework is equipped with four independent agents.
Long Trajectory Learning further grooms the LLM to establish logical associations between agents in an end-to-end manner. 
We train based on the previous stage on the long-trajectory subset $\mathcal{D}_{long}$. 
Specifically, given instruction $x$, long trajectory learning forces the LLM to learn the long trajectory process:
\begin{equation}
\begin{aligned}
\mathcal{L}\left(\mathcal{D}_{Long}\right) &= \sum_{i}  \log P_{L M}\left( \tau_{i}^{R}; \tau_{i}^{I}; \tau_{i}^{G} \mid  x_{i}\right),  \\
\tau _{i}^{T} &= \left [h_{i}^{T};y_{i}^{T};e_{i}^{T}  \right ], T\subset \left \{R,I,G  \right \} .
\end{aligned}
\label{eq:loss3}
\end{equation}
where $R$, $I$ and $G$ denote the Intent Reconstructor, Fact Locator and Response Generator, respectively.
Unlike short trajectory learning (Eq. \ref{eq:loss2}), the framework learns both to predict the target output for each short trajectory as well as from the previous trajectory end $e^{T}$ to the next trajectory head $h^{T+1}$.
In essence, the trajectory token serves as a skeleton in the learning process, guiding the agent not only to grasp a fine-grained representation of the intra-trajectory but also inter-trajectory interactions.

\section{Experiment Setting}
\label{experiment}
\subsection{Setup} 
\paragraph{Task and Dataset.} We evaluate our framework in a range of knowledge-intensive downstream tasks. Including (1) Fact verification: PubHealth \cite{akhtar2022pubhealthtab} is a fact verification dataset about public health; (2) Multiple-choice reasoning: ARC-Challenge \cite{clark2018think} is a multiple-choice questions dataset about science exam. (3) Open-domain question answering: contains two short-form QA datasets, PopQA \cite{mallen2022not}, and SQuAD 1.1 \cite{rajpurkar2016squad}.
(4) Ambiguous question answering: ASQA \cite{gao2023enabling} is ambiguous factoid question of the long form response. Details of evaluation data, including size, and evaluation metrics are available in Appendix Sec. B.1.

\paragraph{Baselines.} We compare our framework with a wide range of baseline methods in two categories. (1) Knowledge internalization methods (General-purpose LLMs): ChatGPT (gpt-3.5-turbo-0125) \cite{zheng2024judging} \cite{ouyang2022training}, Mistral-Instruct-v0.2-7B \cite{jiang2023mistral}, Llama-2-Chat-7B/13B \cite{touvron2023llama}, Vicuna-v1.5-13B \cite{zheng2024judging} and Alpaca2-7B~\footnote{\url{https://github.com/tatsu-lab/stanford_alpaca}} \cite{zheng2024judging}. 
(2) Knowledge enhancement methods: REPLUG-7 \cite{shi2023replug}, VANILLA-7B \cite{gao2023enabling}, INTERACT-7B \cite{gao2023enabling}, RAIT-7B \cite{lin2023ra}, SelfRAG-7B \cite{asai2023self}, MMAgent-3*7B (modular approach).
More details are in Appendix Sec. B.2.

\subsection{Implementation Details} 
\label{sec:Implementation Details}
Due to page limitations, details of our training and evaluation are in Appendix Sec. B.3.

\section{Experiment Result}
\subsection{Main Result}
\paragraph{Comparison against knowledge internalization methods.} 
As shown in Table \ref{tab:main-exp}, our framework shows a significant performance advantage over equivalently sized fine-tuned LLMs across all tasks.
In comparison to larger LLMs (Vicuna-v1.5-13B and Llama-2-Chat-13B), which possess greater internalized knowledge, our SMART framework also exhibits superior performance in all metrics. 
Furthermore, our framework surpasses ChatGPT in all evaluated metrics for PopQA (long-tail knowledge evaluation), Squad1, and ASQA. 
Experimental results indicate that our method more effectively addresses long-tail knowledge, delivering more accurate and fluent responses compared to knowledge internalization methods, which necessitate extensive fine-tuning and training on large volumes of private data.
\begin{table}[]
\centering
\resizebox{0.46\textwidth}{!}{
\begin{tabular}{@{}lllll@{}}
\toprule
             & Health   & ARC-C   & Pop     & AS     \\
             & (Acc)    & (Acc)   & (Acc)   & (Em)   \\ \midrule
\multicolumn{5}{c}{\cellcolor[HTML]{EFEFEF}Training ablation} \\ \midrule
SMART (L)  & \textbf{72.15}  & \textbf{60.22} & \textbf{37.27} & \textbf{36.10}       \\\hdashline
w/o $\mathcal{A}_{\mathrm{f}}$  & 70.13  & 58.95 & 34.31 & 34.77 \\
w/o $\mathcal{A}_{\mathrm{i}}$ & 69.82  & 54.94 & 35.17 & 34.41 \\ 
w/o All    & 57.95  & 56.99 & 21.15 & 20.05 \\\midrule
\multicolumn{5}{c}{\cellcolor[HTML]{EFEFEF}Inference ablation}     \\ \midrule
SMART (L+S) & \textbf{73.18}  & \textbf{65.58} & \textbf{42.60} & \textbf{41.16} \\\hdashline

w/o $\mathcal{A}_{\mathrm{f}}$  & 71.63  & 62.45 & 37.45 & 36.10       \\
w/o $\mathcal{A}_{\mathrm{i}}$ & 71.22  & 60.11 & 39.88 & 35.30     \\
w/o All   & 69.32  & 58.81 & 16.79 & 31.32 \\\bottomrule
\end{tabular}
}
\caption{Training Ablation and inference ablation for the contribution of different agents. L and S denote long-trajectory and short-trajectory learning, respectively. w/o $\mathcal{A}_{\mathrm{f}}$, w/o $\mathcal{A}_{\mathrm{i}}$, and w/o All denote no fact Locator, no intent reconstructor, and only response generator.}
\label{tab: Ablation}
\end{table}

\paragraph{Comparison against knowledge enhancement methods.}
Considering fairness and persuasiveness, we compared knowledge enhancement methods based on the same size as ours.
As shown in Table \ref{tab:main-exp}, our SMART performs better on most tasks compared to other knowledge enhancement methods.
Compared to the SOTA retrieval method, SelfRag \cite{asai2023self}, our model shows great superiority in both accuracy and fluency.
Our method exceeds MMAgent (four independent agents coupled together) in all metrics. This demonstrates that our learning paradigm improves multi-agent collaboration, resulting in more accurate responses. Note that INTERACT \cite{gao2023enabling} is better than us on Squad1, the reason is that INTERACT allows the response model to do more reasoning steps, which is beneficial for hitting answers in short-format generation tasks. RAIT \cite{lin2023ra} is trained with SMART same data and initialized model without fact location and intent reconstruction, lagging behind us.
Overall, our SMART delivers excellent performance in a diverse range of knowledge-intensive tasks.
This result indicates SMART gains are not solely from the multi-agent framework and demonstrate the effectiveness of the long-short trajectory learning.

\subsection{Ablation Studies}

\paragraph{Training ablation of different agents.}
Training ablation aims to verify the superiority of the entire multi-agent combination setup.
To save the experiment cost, we implement long-trajectory learning using 60,000 samples from the long-trajectory subset to evaluate the performance of the co-framework under different agent absence scenarios. As the top part of Table \ref{tab: Ablation}, the absence of the fact Locator and the intent reconstructor significantly degrades the framework's performance. The intent reconstructor provides substantial benefits for multiple-choice reasoning (ARC-C) and ambiguous questions (ASQA), while the fact Locator is crucial for long-tail knowledge Q\&A (PopQA). The experiment proved the effectiveness of different agents in our SMART, especially the fact Locator and the intent reconstructor.

\paragraph{Inference ablation of different agents.} We use the full version of SMART with short long-trajectory learning to ignore the trajectories of different agents during the inference phase. As the bottom part of Table \ref{tab: Ablation}, each agent plays an important role in the collaboration framework. The effect degradation of the fact-checking task (Health) was not severe, which may be related to the large amount of knowledge injected during the short trajectory learning. 
In addition, note that if the inference process is missing a particular agent, most multi-agent frameworks that use end-to-end training become terrible, due to the loss of signals from the missing agent. 
Benefiting from our Short-Trajectory Learning through the trajectory tokens, our SMART does not collapse in performance when an agent is missing, demonstrating flexibility while maintaining performance.

\paragraph{Effects of Long-Short Trajectory Learning.}

\begin{table}[]
\centering
\resizebox{0.46\textwidth}{!}{
\begin{tabular}{@{}lllll@{}}
\toprule
\multirow{2}{*}{Methods} & Health & PopQA & \multicolumn{2}{c}{ASQA} \\ \cmidrule(l){2-5} 
                         & (Acc)  & (Acc) & (Em)        & (R-L)      \\ \midrule
Vanilla LLM              & 9.80    & 22.69 & 14.11       & 6.45       \\
+ Short                  & 62.00  & 32.23 & 23.95       & 19.91      \\
+ Long                   &72.9        & 37.66       &39.86             &39.51            \\ 
+ Short \& Long          & \textbf{73.18}  & \textbf{42.60} & \textbf{41.16}       & \textbf{40.66 }     \\ \bottomrule
\end{tabular}}
\caption{Ablation studies of long-trajectory (Long) and short-trajectory (Short) learning.}
\label{tab:lstL}
\end{table}

\begin{figure*}[t]
    \centering
\includegraphics[width=0.95\linewidth]{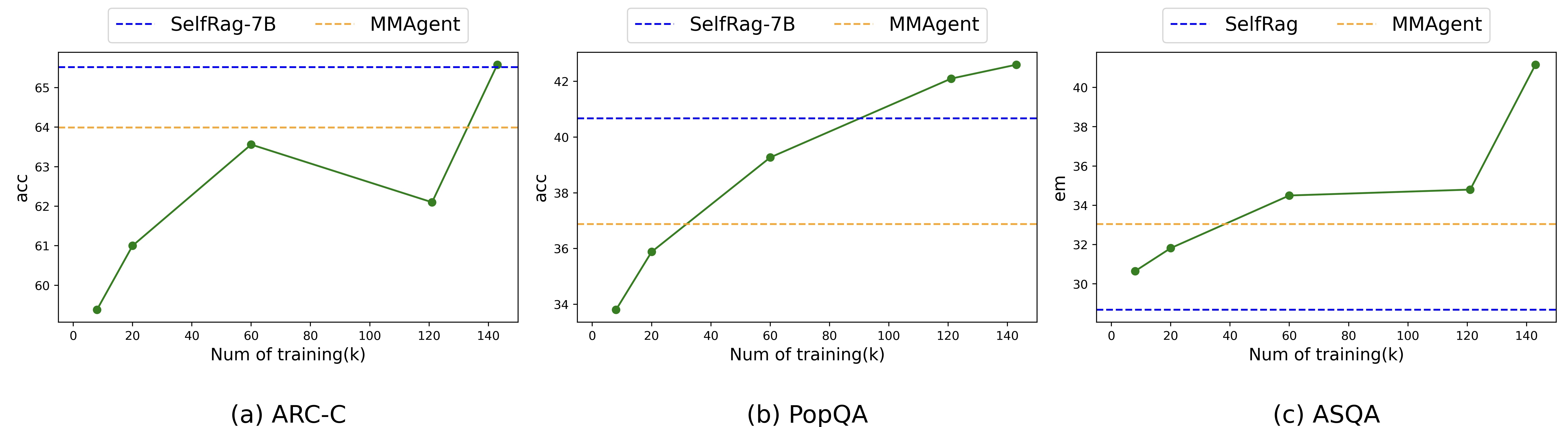}
    \caption{Effects of long-trajectory training data size (K) on three tasks, ARC-C, PopQA and ASQA.}
    \label{fig:Effects_learnig}
\end{figure*}
 Long-Short Trajectory Learning optimising a Multi-agent framework through two-stage learning. we demonstrate its effectiveness progressively by training it on vanilla models, Llama-2-7B-hf \cite{touvron2023llama}. As shown in Table \ref{tab:lstL}, short-trajectory learning and long-trajectory learning enable huge performance improvements in the framework for all tasks. 
Short-trajectory learning enhances the system by optimizing each agent's base capability, though its impact is not as substantial as that of long-trajectory learning.
Long-trajectory learning, by optimizing agent synergy, underscores the importance of collaborative optimization in a multi-agent framework, despite the challenges posed by complex data construction. Overall, the combined approach of long-short trajectory learning yields the best performance, highlighting the significance of simultaneous collaboration and individual uniqueness.

\paragraph{Effects of training data size.}

To examine the impact of long-trajectory training data on long-short trajectory learning, we randomly selected subsets of 8k, 20k, 60k, and 121k instances from the initial 140k
training instances and fine-tuned four SMART variants on these subsets. Subsequently, we compared the model performance on ARC-C, PopQA, and ASQA with our SelfRAG and MMagent models. As shown in Figure \ref{fig:Effects_learnig}, an increase in data size generally leads to improved performance across all datasets. 
Notably, by utilizing 60k data instances, SMART outperformed SelfRAG, which employs 120k samples. This demonstrates the significant advantage of our learning approach in markedly enhancing the performance of multi-agent framework.
\vspace{-2mm}

\section{Related Work}
\label{Work}
\paragraph{Trajectory Learning.} 
Trajectory learning aims to allow agent systems to complete a complex task or scenario through a series of interconnected phases, which requires a profound understanding of both global and local dimensions. Some methods ~\cite{chen2023program,song2023llm,kong2023tptu,asai2023self,sun2022black,mou2024unveiling} enable agent learning trajectory via providing crafted prompt or tuning, which may not consistently yield high performance in every phase. Moreover, independently modules ~\cite{liu2023llm+,shen2024small,ma2023query,xu2023recomp,wang2023learning} can be combined with agent to implement trajectory inference, while this integration confers robust isolated capabilities, the gap between modules might lead to cumulative errors throughout the trajectory process. 
In this paper, we introduce long-short trajectory learning, which equips multi-agent systems with the ability to not only grasp the logic connecting steps but also to refine each step. Our approach is scalable to increasingly complex scenarios.

\paragraph{Knowledge Enhancement Methods.}
Ensuring fact-consistent responses is a core goal of intelligent systems research~\cite{wang2022logic,tu2024context,tu2024distractors,tu2023self,yue2024lawllm,yue2023i3n,gao2024fine}.
LLMs parameterize knowledge by training on gargantuan textual corpora.
However, LLMs suffer from hallucination \cite{ji2023survey}, trouble in acquiring long-tailed fact \cite{kandpal2023large} and struggle to expand their parametric knowledge. For knowledge-intensive scenarios, existing methods~\cite{izacard2023atlas,sun2020colake} usually assist LLMs by integrating non-parametric knowledge. 
Recent advances incorporated retrievers ~\cite{asai2023self,shi2023replug,lin2023ra} to augment LLMs.
The efficacy of non-parametric knowledge collaboration in improving task performance significantly relies on the relevance of the acquired knowledge and the level of knowledge utilization by the LLM itself. However, existing work has not comprehensively confronted these challenges
Some works~\cite{xu2023recomp,ma2023query} simply select relevant knowledge and demonstrate better intentions by combining separate modules.
Self-RAG~\cite{asai2023self} integrates specialized feedback tokens into the language model to assess the necessity for retrieval and to verify the relevance, support, or completeness of the output.
Unlike existing approaches, we introduce a novel multi-agent framework that addresses these challenges with trajectory learning.

\section{Conclusions}

In this paper, we introduce SMART, a novel multi-agent framework that addresses the challenges of generating factually consistent responses in knowledge-intensive tasks. By leveraging external knowledge and employing specialized agents, SMART enhances the interpretability and factual consistency of LLMs generated responses. Our proposed Long- and Short-Trajectory Learning paradigm ensures synergistic collaboration among agents while maintaining fine-grained execution, enabling the framework to navigate complex knowledge-intensive tasks effectively. Empirical results on five diverse tasks demonstrate SMART's superior performance compared to SOTA pre-trained and instruction-tuned LLMs, as well as widely adopted methods.  SMART highlights the importance of integrating external knowledge and employing multi-agent systems to tackle the limitations of LLMs in knowledge-intensive scenarios.   

\textbf{Future work}. One is that our framework currently executes sequentially without iterative optimization, which may lead to insufficient knowledge retrieval for multi-hop problems. However, this can be addressed by adding loop arrows between the Fact Locator and Intent Reconstructor agents. Another is that our retriever is not trained in the whole process, although it can be incorporated into the training process using existing techniques. We envision our framework as a general paradigm that extends beyond knowledge-intensive tasks to more complex scenarios, enabling any multi-agent framework to internalize tailored trajectories.

\section{Acknowledgments}
This research is supported by National Key R\&D Program of China (2023YFF1204800) and National Natural Science Foundation of China (No.62176058). The project’s computational resources are supported by CFFF platform of Fudan University.

\bibliography{aaai25}

\newpage


\clearpage
\appendix
\section{A\hspace{0.2cm} Data Construction}
\subsection{A.1\hspace{0.2cm} Full list of datasets.} 
\label{sec:appendix_data}
\begin{table*}[]
\centering
\setlength{\tabcolsep}{3.6mm}{
\begin{tabular}{@{}llccccc@{}}
\toprule
\multirow{3}{*}{\textbf{Category}}                                                        & \multirow{3}{*}{\textbf{Dataset Name}} & \multirow{3}{*}{\textbf{Long-Trajectory}} & \multicolumn{4}{c}{\textbf{Short-Trajectory}}             \\ \cmidrule(l){4-7} 
                                                                                 &                               &                                  & Loctor   & Intent & \multicolumn{2}{c}{Response} \\ \cmidrule(l){6-7} 
                                                                                 &                               &                                  &          &        & w/o $\mathcal{F}$        & w/ $\mathcal{F}$         \\ \midrule
Fact Verification                                                                & FEVER                         & 22,485                           & 25,211   & 10,761 & 1,757         & 3,513        \\ \midrule
\multirow{2}{*}{Dialogue}                                                        & QReCC                         & 18,363                           & -        & 21,000 & 3,383         & -            \\
                                                                                 & WoW                           & 10,701                           & -        & -      & 3,824         & -            \\ \midrule
\multirow{5}{*}{Open-domain Q\&A}                                                & Natural Questions             & 17,770                           & 25,065   & 15,069 & 1,347         & 2,695        \\
                                                                                 & HotpotQA                      & 10,792                           & 25,325   & 10,200 & 1,356         & 2,714        \\
                                                                                 & 2WikiMHQA                     & 9,847                            & 25,118   & 17,140 & 1,675         & 3,348        \\
                                                                                 & StrategyQA                    & 2,281                            & -        & -      & -             & -            \\
                                                                                 & ASQA                          & 4,236                            & -        & -      & 667           & 1,335        \\ \midrule
\multirow{3}{*}{\begin{tabular}[c]{@{}l@{}}Commonsense\\ Reasoning\end{tabular}} & CosmosQA                      & 6,923                            & 6,315    & -      & -             & 2,526        \\
                                                                                 & OpenbookQA                    & 4,923                            & -        & -      & 495           & 992          \\
                                                                                 & Arc-Easy                      & 2,242                            & -        & -      & -             & -            \\
                                                                                 & ThoughtSource                 & 5,918                            & -        & 20,000 & 18,711        & -            \\ \midrule
\multirow{5}{*}{General}                                                         & Alpaca-GPT4                   & 13,944                           & -        & -      & 26,001        & -            \\
                                                                                 & Alpaca-Stanford               & 12,082                           & -        & -      & 25,989        & -            \\
                                                                                 & FLAN-V2                       & -                                & -        & -      & 30,316        & -            \\
                                                                                 & Dolly                         & -                                & -        & -      & 6,802         & -            \\
                                                                                 & OpenAssistant                 & -                                & -        & -      & 19,141        & -            \\ \midrule
Total                                                                            &                               & 142,507                          & 107,034  & 94,170 & 141,464       & 17,123       \\ \bottomrule
\end{tabular}}
\caption{Distribution of our dataset. w/o $\mathcal{F}$ indicates that  response generator without facts guidance, w/ $\mathcal{F}$ on the contrary}
\label{tab:dataset}
\end{table*}

For coverage of a wide range of data sources, we collected instances from four categories of knowledge-intensive scenarios: 
(1) Fact Checking includes FEVER \cite{thorne2018fever}. 
(2) Dialogue includes Wizard of Wikipedia \cite{dinan2018wizard} and QReCC \cite{anantha2021open}. (3) Open-domain Q\&A includes Natural Questions \cite{kwiatkowski2019natural}, HotpotQA \cite{yang2018hotpotqa}, 2WikiMHQA \cite{ho2020constructing}, 
StrategyQA \cite{geva2021did} and ASQA \cite{stelmakh2022asqa}. 
(4) Commonsense Reasoning includes ComsmosQA \cite{huang2019cosmos}, ARC-Easy \cite{clark2018think}, OpenBookQA \cite{mihaylov2018can} and ThoughtSource \cite{wei2022chain}. 
In addition, we also sampled from some generic instruction datasets to ensure flexibility and creativity in our framework, including GPT-4 Alpaca \cite{peng2023instruction}, Alpaca~\footnote{\url{https://github.com/tatsu-lab/stanford_alpaca}}, OpenAssistant \cite{kopf2304openassistant}, FLAN-V2 \cite{longpre2023flan}, and Dolly ~\footnote{\url{https://huggingface.co/datasets/databricks/databricks-dolly-15k}}. Table \ref{tab:dataset} shows the full list of training instances. The total dataset amount is 500k instances, where the long-trajectory subset includes 140k well-designed instances, and the short-trajectory subset contains 360k easily accessible instances.

\subsection{A.2\hspace{0.2cm} Datasets Construction Details} 
\label{sec:appendix_data_detail}


We propose two distinct types of datasets: Long-trajectory Datasets and Short-trajectory Datasets, which apply in different stages of long short-trajectory learning, respectively. These datasets differ in their structure, objectives, and the way they are used to train models, as shown in Table~\ref{tab:dataset}. By leveraging both types of datasets, we can develop robust and versatile framework that combine the benefits of task-specific training with the power of end-to-end reasoning.

\paragraph{Long-trajectory Datasets.}{We use the following steps to construct the long trajectory dataset.

\textbf{Step1}. As shown in Table \ref{tab:dataset}, we leverage existing datasets and transform them into a unified format. Let $\mathcal{D} = {(x_i, y_i)}_{i=1}^N$ denote a QA dataset, where $x_i$ represents the question and $y_i$ represents the corresponding answer. For multi-turn dialogue datasets, we concatenate the historical context and the current question to form $x_i$, and use the answer from the last turn as $y_i$. This allows us to standardize both single-turn datasets and multi-turn dialogue datasets into a consistent format of $(x_i, y_i)$ pairs.

\textbf{Step2.} To generate the Long-trajectory Datasets, we employ GPT-4 and an off-the-shelf retriever. For each input $x$, we use few-shot prompting to generate multiple query texts $(q_1, q_2, ..., q_m)$, where the number of queries $m$ is determined by GPT-4. The prompting strategies are tailored to different types of datasets, including Fact Verification, Multi-turn Dialogue, Open-domain Q\&A, Commonsense Reasoning, and etc.

\textbf{Step3.} For each query text $q_j$ $(j = 1, 2, ..., m)$, we retrieve $k$ candidate passages from a fixed knowledge base (e.g., Wikipedia), resulting in a total of $m \times k$ candidate passages. 

\textbf{Step4.} We utilize few-shot prompting with GPT-4 to determine the relevance of each candidate passage in answering the user's question. Specifically, for each passage, GPT-4 is prompted to identify the specific facts (multiple sentences in the the passage) that can help answer the question, or output "irrelevant" if the passage does not contain relevant information. We require that the facts outputted by GPT-4 must be contained within the passage. 

\textbf{Step5.} The final answer consists of the original answer and the passage number of the facts supporting the answer. The Long-trajectory Datasets are constructed by combining the reconstructed queries, retrieved passages, located facts, and generated answer into a single sequence. For each $(x, y)$ pair, we transform $y$ into a long-trajectory reasoning process as follows:

\begin{verbatim}
Input: x
------------------------------------
Output:  
<Reconstructor> q1, q2, ... </eor>
<retrieval> 
[1] xxxxxxxx 
[2] xxxxxxxx 
[3] xxxxxxxx 
...
</retrieval>
<Locator> 
[Relevant]:[1] xxxx
[Irrelevant]:[2] Lacking Supporting Facts.  
[Irrelevant]:[3] Lacking Supporting Facts. 
...
</eol>
<Generator> y [Cite]: [1]</eog>
\end{verbatim}

By reconstructing the original answer $y_i$ into a long-trajectory reasoning process, we explicitly model the steps of query rewriting, retrieval, fact locating, and question answering. This approach allows us to create datasets that showcase the complex reasoning capabilities required for knowledge-intensive tasks, providing valuable insights and resources for advancing research in this area. The loss function is computed only for the components generated by the large language model, i.e., the \raisebox{-0.2em}{\includegraphics[height=1em]{intent.pdf}}, \raisebox{-0.2em}{\includegraphics[height=1em]{loctar.pdf}}, and \raisebox{-0.2em}{\includegraphics[height=1em]{generator.pdf}} sections, while the \raisebox{-0.2em}{\includegraphics[height=1em]{rhead.pdf}} section is excluded from the loss computation. }


\paragraph{Short-trajectory Dataset.}
In contrast to the Long-trajectory Datasets, the Short-trajectory Datasets focus on the individual characteristics of each agent. These datasets are constructed to target specific skills, such as intent reconstruction, fact location, and response generation, and can be used for pre-training models or training specialized agents. This property allows such dataset to be obtained directly from existing open source datasets, as shown in Table \ref{tab:dataset}.

\textbf{For intent reconstruction}. The construction of Short-trajectory Datasets follows the same approach as Long-trajectory Datasets. These datasets can be formalized as:

\begin{verbatim}
Input: x 
<Reconstructor> 
------------------------------------
Output:  q1, q2, ...  </eor>
\end{verbatim}

where $q_i$ represents the reconstructed knowledge query intent from the user input $x$.

\textbf{For fact location}. We can leverage existing datasets such as HotpotQA, Natural Questions, and 2WikiMHQA. Fact location does not require the output from the reconstructor, instead, it only needs the user input and the retrieved passages. These datasets can be formalized as:

\begin{verbatim}
Input: x
<retrieval> 
[1] xxxxxxxx 
[2] xxxxxxxx 
[3] xxxxxxxx 
</retrieval>
<Locator> 
------------------------------------
Output:  
[Relevant]:[1] xxxx
[Irrelevant]:[2] Lacking Supporting Facts.  
[Irrelevant]:[3] Lacking Supporting Facts. 
<eol> 
\end{verbatim}

\textbf{For response generation}. We construct two types of training samples, one type that directly generates an answer based on the user question: 

\begin{verbatim}
Input: x
<Generator> 
------------------------------------
Output: y </eog>
\end{verbatim}

Another is that answers the question based on both the question and the located facts: 

\begin{verbatim}
Input: x
<Locator> 
[Relevant]:[1] xxxx
[Irrelevant]:[2] Lacking Supporting Facts.  
[Irrelevant]:[3] Lacking Supporting Facts. 
</eol>
<Generator> 
------------------------------------
Output: y [Cite]: [1] </eog>
\end{verbatim}

The Short-trajectory Datasets offer several advantages. First, they do not necessarily require a complete long-trajectory training dataset, allowing us to utilize a large number of existing NLP datasets. Second, they enable focused training on individual skills, and subsequent short trajectories do not need to depend on the outputs of all previous short trajectories. By pre-training LLMs on cost-effective Short-trajectory Datasets, we can reduce the amount of cost-ineffective Long-trajectory training data required to achieve better performance comparable to LLMs without pre-training. 

Furthermore, the Short-trajectory Datasets can be used to train specialized agents, each responsible for a specific task (e.g., intent reconstruction, fact location, or response generation). These agents can then be combined to form a complete question-answering system, offering a modular and adaptable approach to solving complex tasks.



\subsection{A.3\hspace{0.2cm} Datasets example}
Table \ref{tab:examplse_long_training_table} 
shows two training cases in the long-trajectory subset
and Table \ref{tab:examplse_short_training_table} shows the four training
cases in the short-trajectory subset.

\section{B\hspace{0.2cm} Experimental Setups}
\label{sec:appendix}

\subsection{B.1\hspace{0.2cm} Dataset and Evaluation Metrics}
\label{sec:appendixdata}
\begin{itemize}
    \item \textbf{Fact verification}: PubHealth \cite{akhtar2022pubhealthtab} is a fact verification dataset about public health. We use accuracy as an evaluation metric and report on the test set, which contains 987 test examples with veracity labels (true, false).
    \item \textbf{Multiple-choice reasoning}: ARC-Challenge \cite{clark2018think} is a multiple-choice questions dataset about science exams, containing 1172 test examples. We also use accuracy as an evaluation metric.
    \item \textbf{Open-domain question answering}: 1) PopQA \cite{mallen2022not} is a long-tailed set of 1,399 rare entity queries collected from Wikipedia.
    2) SQuAD 1.1 \cite{rajpurkar2016squad} contains 8,886 queries that is created through a process where annotators write questions based on the documents they read.
    Following previous practice \cite{asai2023self,gao2023enabling}, we assessed this track accuracy by matching, \textit{i.e.}, ground truth appears in the model responses.
    \item \textbf{Ambiguous question answering}: ASQA \cite{gao2023enabling}  containsis 4132 ambiguous factoid questions of the long-form response.  Following the official setting \cite{gao2023enabling}, we use Mauve to assess fluency, and Str\_EM and Rouge-L (R-L) to assess accuracy. 
\end{itemize}

\subsection{B.2\hspace{0.2cm} Baselines}
\label{sec:Baselines}
\paragraph{Knowledge internalization methods:}
LLMs internalize a great deal of world knowledge in their parameters, so generic LLMs are regarded as Knowledge internalization methods.
In training and inference, we use the official system prompt or instruction format. knowledge internalization methods are as follows:
\begin{itemize}
    \item Instruct fine-tuned and preference alignment models: ChatGPT~\footnote{We use gpt-3.5-turbo-0125.} \cite{ouyang2022training}, Llama-2-Chat-7B \cite{touvron2023llama}, Llama-2-Chat-13B \cite{touvron2023llama}.
    \item Instruct fine-tuned models: Instruct-v0.2-7B \cite{jiang2023mistral}, Vicuna-v1.5-13B \cite{zheng2024judging}, and Alpaca2-7B (Following Alpaca~\footnote{\url{https://github.com/tatsu-lab/stanford_alpaca}}, we trained based on Llama-2). 
\end{itemize}
 
\begin{table}[]
\centering
\resizebox{\columnwidth}{!}{%
\begin{tabular}{ll}
\hline
\textbf{Dataset} & \multicolumn{1}{c}{\textbf{Instruction}}                                                                                                                                                                                                    \\ \hline
ARC-C            & \begin{tabular}[c]{@{}l@{}}Given four answer candidates, choose \\ the best answer choice.\end{tabular}                                                                                                                                     \\ \hline
PubHealth        & \begin{tabular}[c]{@{}l@{}}Is the following statement correct\\  or not? Say true if it's correct; otherwise, \\ say false.\end{tabular}                                                                                                    \\ \hline
ASQA             & \begin{tabular}[c]{@{}l@{}}Answer the following question. The \\ question may be ambiguous and have \\ multiple correct answers, and in that case, \\ you have to provide a long-form answer \\ including all correct answers.\end{tabular} \\ \hline
\end{tabular}%
}
\caption{Instructions for zero-shot evaluations.}
\label{tab:taskinstuction}
\end{table}
\paragraph{Knowledge enhancement methods:}
We employ widely used knowledge augmentation methods. Since some of the methods do not provide model weights, we replicated them using the same base model and data as ours. In addition, we also use the same retrieval model and knowledge base as ours for fairness. Knowledge enhancement methods  are as follows:
\begin{itemize}
    \item \textbf{REPLUG} \cite{shi2023replug} treats the frozen black-box LLM and augments it with a tuneable retrieval model. We use Llama-2-Chat-7B \cite{touvron2023llama} as black-box LLM.
    \item \textbf{VANILLA-7B} \cite{gao2023enabling} is a framework that retrieves passages, then instructs the model to distinguish relevant documents and cite accordingly. We use Llama-2-Chat-7B as the backbone.
    \item \textbf{INTERACT-7B} \cite{gao2023enabling} is an interactive prompting scheme that allows the agent to check passages by executing the "Check" "Output" and "End" actions. We use Llama-2-Chat-7B as the backbone.
    \item \textbf{RAIT-7B} \cite{lin2023ra} retrofit LLMs with retrieval capabilities by tuning. To be fair, we train pre-trained Llama-2 using the same dataset as ours.
    \item \textbf{SelfRAG-7B} \cite{asai2023self} is a framework to enhance the response quality through retrieval on demand and self-reflection. SelfRAG uses an end-to-end optimization strategy.
    \item \textbf{MMAgent-3*7B} is our setting modular multi-agent framework. We train separate agents based on the same dataset and complete the workflow by decoupling them. We use same pre-trained Llama-2 as backbone to train three independent agents.
\end{itemize}

\subsection{B.3\hspace{0.2cm} Setting Details}
\label{sec:trainingdetails}

\paragraph{Training Detail.} 
    We use pre-trained LLM, Llama2-7B \cite{touvron2023llama}, as our initial model. We use 8*V100 GPUs with 32GB memory to conduct our Short-Long Trajectory learning by LoRA method \cite{hu2021lora}.
    Both short and long-trajectory learning are conducted over 2 epochs with a total batch size of 128, a peak learning rate of 2e-4, and 3\% warmup steps, followed by linear decay. The maximum token length is set to 3,076 for short-trajectory learning and 2,816 for long-trajectory learning. Multi-GPU distributed training is performed using DeepSpeed Stage 3 \cite{rasley2020deepspeed}. 


\paragraph{Evaluation Details.}
Knowledge Retriever is driven by Retriever-MSMARCO \cite{izacard2021unsupervised} and access top-3 knowledge documents from the official Wikipedia corpus \cite{karpukhin2020dense}. These documents are segmented into non-overlapping fragments of 100 words.  
In the evaluation, we conduct zero-shot assessment, \textit{i.e.}, and we provide instructions describing tasks without few-shot demonstrations \cite{asai2023self}. Greedy decoding was used across the evaluations.

\paragraph{Evaluation Task Instructions.} 
Following the existing work \cite{asai2023self}, in addition to open-domain Q\&A, we implemented zero-shot evaluations by providing prompts for task descriptions as shown in Table \ref{tab:taskinstuction}.  

\section{C\hspace{0.2cm} Inference} 
\paragraph{Inference Overview.} Algorithm \ref{algo:inference} gives an inference overview of our multi-agent framework. The systematic procedure is delineated in the following steps:
$\mathcal{A}_{\mathrm{i} }$  first mines the explicit intent $\bar{q} = \left \{ q_{1},q_{2},...,q_{m}  \right \}$ from the instruction $x$.
Next, $\mathcal{A}_{\mathrm{r} }$ retrieves top-k knowledge documents $\bar{d}=\left \{  d_{1},
d_{2},...,d_{k\times m} \right \} $ using each intent.
Then, $\mathcal{A}_{\mathrm{l} }$ determines each relevant knowledge passage and further locates the fact span $f \subset d_{k\times m}$. Finally, $\mathcal{A}_{\mathrm{g} }$ utilizes the previous execution trajectory to generate response $y$ and citations when facts exist, otherwise $\mathcal{A}_{\mathrm{g}}$ utilizes only $x$. 

\begin{algorithm}[tb]
\caption{ Inference}
\label{algo:inference}
\textbf{Require}: Intent Reconstructor 
$\mathcal{A}_{\mathrm{i} }$, Knowledge Retriever $\mathcal{A}_{\mathrm{r} }$, Fact Locator $\mathcal{A}_{\mathrm{l} }$, and Response Generator $\mathcal{A}_{\mathrm{g} }$, Passage collections ${d_{1}, . . . , d_{k*m}}$, trajectory head token $h$, trajectory end token $e$\\
\textbf{Input:} user prompt $x$ \\  \textbf{Output:} $y$
\begin{algorithmic}[1]
\STATE  $\mathcal{A}_{\mathrm{i}}$ predicts ${q_{1}, . . . , q_{m}}$,$e_{i}$,$h_{r}$ given $x$,$h_{i}$
\FOR{each $p$ in ${q_1, . . ., q_m}$}
    \STATE Retrieve passages ${d_1, . . ., d_k}$ using $\mathcal{A}_{\mathrm{r}}$ given $p$, top-$k$
\ENDFOR
\STATE $q = \{d_1, . . ., d_{k \cdot m}\}$
\STATE $\mathcal{A}_{\mathrm{l} }$ predicts $\{(r_1, f_1), . . ., (r_{k \cdot m}, f_{k \cdot m})\}$, $e_l$, $h_g$ given $x$, $\{d_1, . . ., d_{k \cdot m}\}$, $e_r$, $h_l$
\IF{$r = \text{[Relevant]}$}
\STATE $\mathcal{A}_{\mathrm{g} }$ predicts $y$, $e_g$ given $\{(r_1, f_1), . . ., (r_{k \cdot m}, f_{k \cdot m})\}$, $e_l$, $h_g$
\ELSE
\STATE $\mathcal{A}_{\mathrm{g} }$ predicts $y$, $e_g$ given $x$, $h_g$
\ENDIF
\end{algorithmic}
\end{algorithm}

\section{D\hspace{0.2cm} Case Study} 
\begin{figure*}
    \centering
\includegraphics[width=0.9\linewidth]{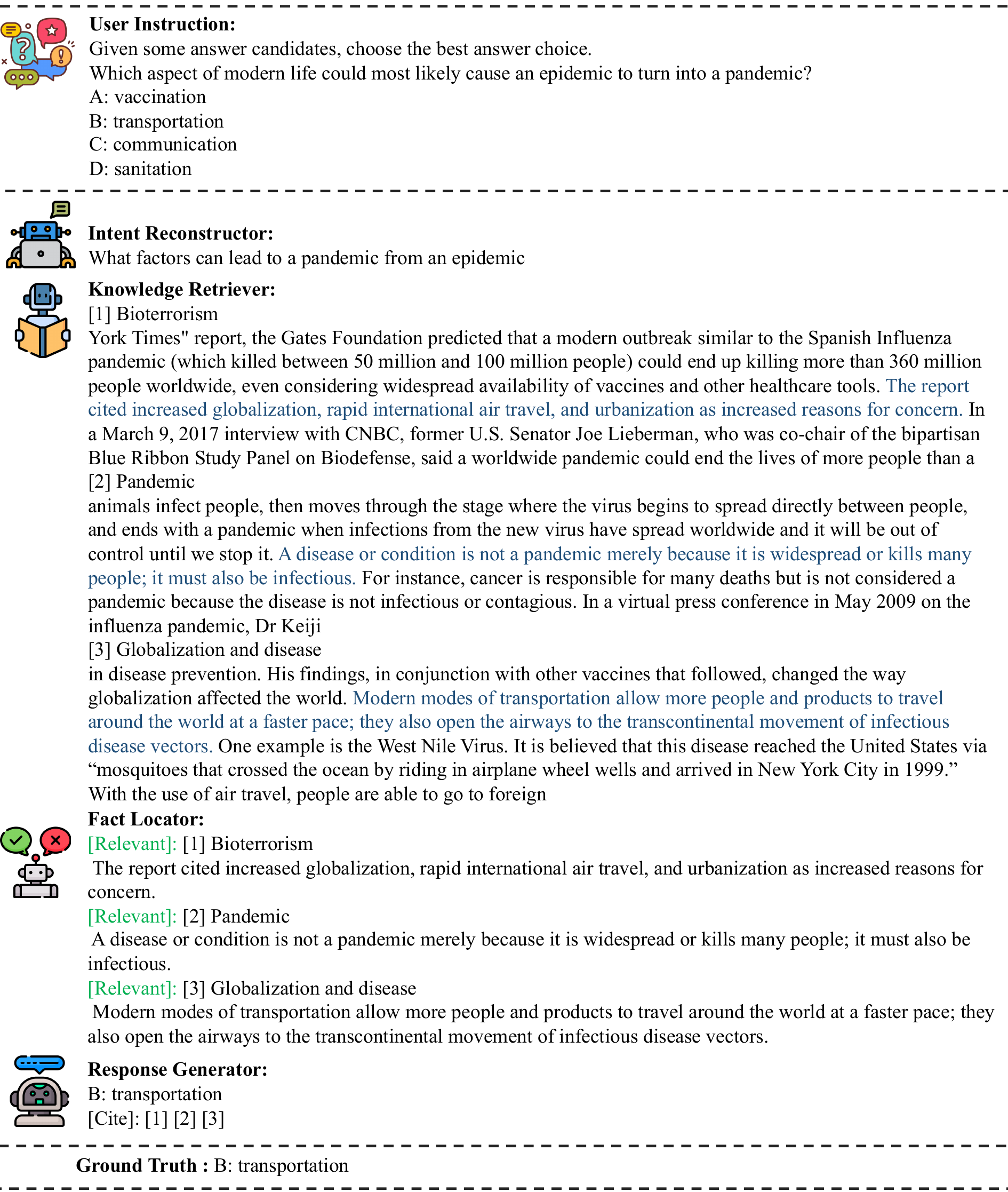}
    \caption{Example of our SMART output on ARC-Challenge}
    \label{fig:case1}
\end{figure*}

\begin{figure*}
    \centering
\includegraphics[width=0.88\linewidth]{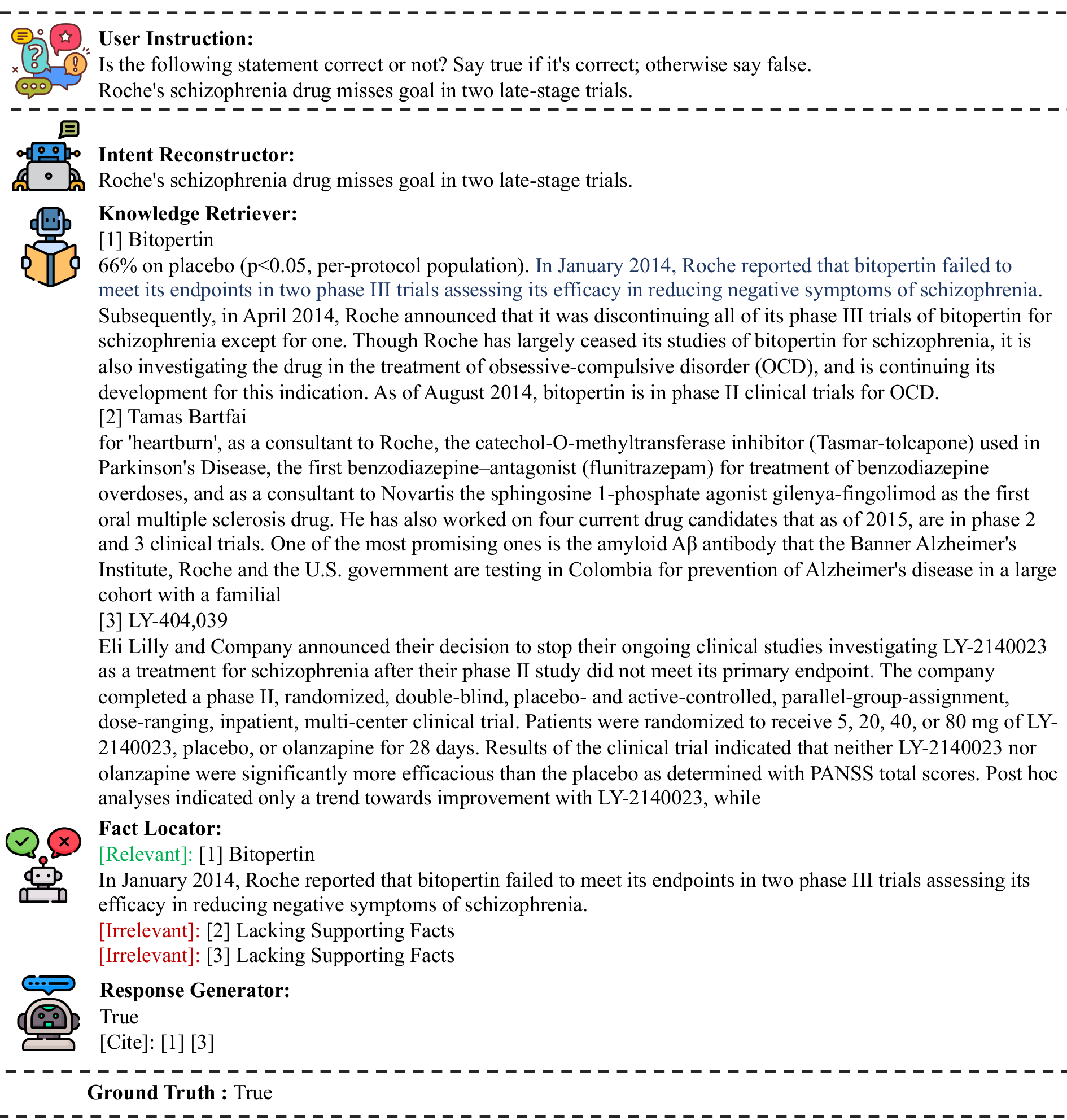}
    \caption{Example of our SMART output on PubHealth}
    \label{fig:case2}
\end{figure*}

\begin{figure*}
    \centering
\includegraphics[width=0.9\linewidth]{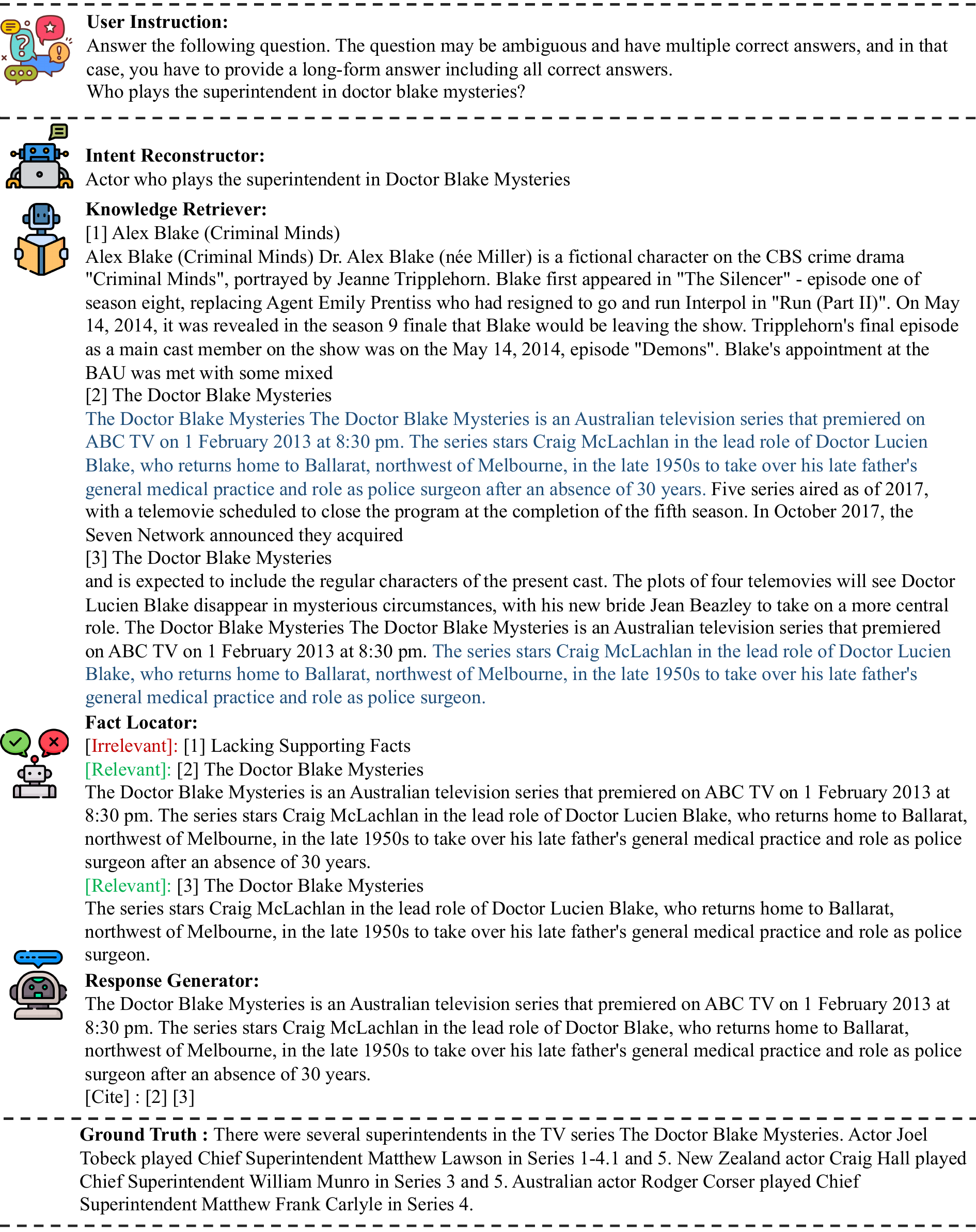}
    \caption{Example of our SMART output on ASQA}
    \label{fig:case3}
\end{figure*}

To present the performance generated by the proposed SMART, we conduct case studies on ARC-Challenge \cite{clark2018think}, ASQA \cite{gao2023enabling}
and PubHealt \cite{akhtar2022pubhealthtab} datasets.
In the short-form generation, as shown in Figure \ref{fig:case1}, the Intent Reconstructor clarifies the instruction of multiple-choice reasoning into formally and semantically clear knowledge intent. 
Figure \ref{fig:case2} shows that the Fact Locator can correctly evaluate Document 3 as [Irrelevant], which is prone to misidentification.
In the long-form generation, as shown in Figure \ref{fig:case3}, we can observe that Response Generator generates a more fluent instruction response even compared to Ground Truth. Overall, our multi-agent framework can effectively solve different types of knowledge-intensive tasks through reconstructing knowledge intent, accessing external knowledge, discriminating relevant information to identify factual spans, and ultimately crafting informed responses. This process enhances the knowledge interpretability and response factuality, which demonstrates the effectiveness and advancement of our framework.


\begin{table*}[t!]
\renewcommand{\arraystretch}{1.2}
\setlength{\tabcolsep}{2pt}
\footnotesize
    \centering
    \begin{tabular}{p{14cm}}
\toprule
{\bf $<$Instruction$>$} when was alice through the looking glass filmed? (from \texttt{nq})
  \vspace{0.2em}
           \\\hdashline
{\bf $<$Reconstructor$>$} Search(Key figures in the War of 1812){\texttt{$<$/eor$>$}} \\ 
{\texttt{$<$retrieval$>$}}{[1]} War of 1812 -War of 1812 The War of 1812 was a conflict fought between the United States, the United Kingdom, and their respective allies from June 1812 to February 1815. Historians in Britain often see it as a minor theater of the Napoleonic Wars; in the United States and Canada, it is seen as a war in its own right, though it was separate. From the outbreak of war with Napoleonic France, Britain had enforced a naval blockade to choke off neutral trade to France, which the US contested as illegal under international law. To man the blockade, Britain impressed American merchant\\
{[2]} War of 1812 -to the United States. Some British officers and Canadians objected to handing back Prairie du Chien and especially Mackinac under the terms of the Treaty of Ghent. However, the Americans retained the captured post at Fort Malden, near Amherstburg, until the British complied with the treaty. Fighting between Americans, the Sauk, and other indigenous tribes continued through 1817, well after the war ended in the east. In 1812, Britain's Royal Navy was the world's largest, with over 600 cruisers in commission and some smaller vessels. Although most of these were involved in blockading the French navy and protecting British trade\\
{[3]} War of 1812 -War of 1812", sees the British as having fought to a much stronger position than the United States. Even tied down by ongoing wars with Napoleonic France, the British had enough capable officers, well-trained men, and equipment to easily defeat a series of American invasions of Canada. In fact, in the opening salvos of the war, the American forces invading Upper Canada were pushed so far back that they ended up surrendering Michigan Territory. The difference between the two navies was even greater. While the Americans famously (shockingly for contemporaries on both sides of the Atlantic) bested British ships in {\texttt{$<$/retrieval$>$}} \\
{\bf $<$Locator$>$}
{[\texttt{Relevant}]:} {[1]} War of 1812 -The War of 1812 was a conflict fought between the United States, the United Kingdom, and their respective allies from June 1812 to February 1815.\\
{[\texttt{Relevant}]:} {[2]} War of 1812 -Fighting between Americans, the Sauk, and other indigenous tribes continued through 1817, well after the war ended in the east.\\
{[\texttt{Relevant}]:} {[3]} War of 1812 -War of 1812", sees the British as having fought to a much stronger position than the United States.{\texttt{$<$/eol$>$}}\\
{\bf $<$Generator$>$} the United States , the United Kingdom , and their respective allies \\
{[Cite]:} {[1]} {[2]} {[3]}{\texttt{$<$/eog$>$}}
\\
\midrule
{\bf $<$Instruction$>$} Lichens are symbiotic organisms made of green algae and fungi. What do the green algae supply to the fungi in this symbiotic relationship? -A: carbon dioxide -B: food -C: protection -D: water{\texttt{$<$/eoi$>$}} (from \texttt{openbook})  \vspace{0.2em}
           \\\hdashline
{\bf $<$Reconstructor$>$} Search(Symbiotic relationship between lichens -What do green algae supply to fungi in a lichen symbiotic relationship){\texttt{$<$/eor$>$}} \\ 
{\texttt{$<$/retrieval$>$}} {[1]} Symbiosis in lichens -a very different form (morphology), physiology, and biochemistry than any of the constituent species growing separately. The algae or cyanobacteria benefit their fungal partner by producing organic carbon compounds through photosynthesis. In return, the fungal partner benefits the algae or cyanobacteria by protecting them from the environment by its filaments, which also gather moisture and nutrients from the environment, and (usually) provide an anchor to it. The majority of the lichens contain eukaryotic autotrophs belonging to the Chlorophyta (green algae) or to the Xanthophyta (yellow-green algae). About 90\% of all known lichens have a green alga as a symbiont. Among\\
{[2]} .....\\
{[3]} Cyanobacteria -"Chara", where they may fix nitrogen. Cyanobacteria such as "Anabaena" (a symbiont of the aquatic fern "Azolla"), can provide rice plantations with biofertilizer. Many cyanobacteria form motile filaments of cells, called hormogonia, that travel away from the main biomass to bud and form new colonies elsewhere. The cells in a hormogonium are often thinner than in the vegetative state, and the cells on either end of the motile chain may be tapered. To break away from the parent colony, a hormogonium often must tear apart a weaker cell in a filament, called a necridium. Each individual cell (each single cyanobacterium){\texttt{$<$/retrieval$>$}} \\
{\bf $<$Locator$>$}\\
{[\texttt{Relevant}]:}{[1]} Symbiosis in lichens -The algae or cyanobacteria benefit their fungal partner by producing organic carbon compounds through photosynthesis.\\
{[\texttt{Relevant}]:} {[2]} Algae -In these symbioses, the algae supply photosynthates (organic substances) to the host organism providing protection to the algal cells. The host organism derives some or all of its energy requirements from the algae.\\
{[\texttt{Irrelevant}]:} {[3]} Lacking Supporting Facts.{\texttt{$<$/eol$>$}}\\
{\bf $<$Generator$>$} B: food \\
{[Cite]:} {[1]} {[2]}{\texttt{$<$/eog$>$}}\\
\bottomrule
 \end{tabular}
    \caption{Training examples of the long-trajectory subset.}
    \label{tab:examplse_long_training_table}

\end{table*}

\begin{table*}[t!]
\renewcommand{\arraystretch}{1.2}
\setlength{\tabcolsep}{2pt}
\footnotesize
    \centering
    \begin{tabular}{p{14cm}}
\toprule
{\bf $<$Instruction$>$} What did Rahul Bose do in 1993?
-Rahul Bose starred in the 1993 Are There Tigers in the Congo.
What was Rahul Bose's early career? 
-Bose started his acting career on the Mumbai stage in Rahul D'Cunha's Topsy Turvey and Are There Tigers In The Congo?
What other films has Rahul Bose starred in besides Topsy Turvey and Are There Tigers In The Congo?
-D'Cunha's aunt was the casting director for director Dev Benegal's film English, August and suggested that Rahul Bose should play the lead role.
What else has Rahul Bose done in his career besides Topsy Turvey, Are There Tigers in The Congo and English, August?
-After English, August, Rahul Bose found work in television
What was the name of the show Rahul Bose found work in after English, August?
-Rahul Bose was offered a role in India's first English-language television serial, A Mouthful of Sky
-What other TV series has Rahul Bose starred in besides A Mouthful of Sky?
-Rahul Bose also co-hosted BBC World's Style! with Laila Rouass.
When was Rahul Bose on BBC besides BBC World's Style! with Laila Rouass?
-In 1997, Bose was cast to play the role of Saleem Sinai in the BBC adaptation of Salman Rushdie's novel Midnight's Children.
What did he do in 2003?{\texttt{$<$/eoi$>$}}\\
{\bf $<$Reconstructor$>$} (from \texttt{Intent Reconstructor: qrec})
\vspace{0.2em}
           \\\hdashline
Search(Current status of Grey's Anatomy TV show){\texttt{$<$/eor$>$}} 
\\
\midrule
{\bf $<$Instruction$>$} Name three biometrics technologies.{\texttt{$<$/eoi$>$}}  \\
{\bf $<$Generator$>$} (from \texttt{Response Generator: alpaca})
\vspace{0.2em}
           \\\hdashline
1) Fingerprint recognition: This is a technology that identifies individuals by analyzing their unique fingerprint patterns. 2) Face recognition: This is a biometric technology that analyzes the facial features of an individual in order to identify or verify their identity. 3) Retinal scanning: This is a biometric technology that uses the unique pattern of blood vessels in an individual's retina in order to identify them.{\texttt{$<$/eog$>$}}
\\
\midrule
{\bf $<$Instruction$>$} Which film was released more recently, Kora Terry or Yi Yi?{\texttt{$<$/eoi$>$}} \\
{\bf $<$Locator$>$}
-{[\texttt{Irrelevant}]:} {[1]} Lacking Supporting Facts 
-{[\texttt{Relevant}]:}  {[2]} Kora Terry -Kora Terry is a 1940 German drama film directed by Georg Jacoby and starring Marika Rökk, Will Quadflieg and Josef Sieber.
-{[\texttt{Irrelevant}]:} {[3]} Lacking Supporting Facts
-{[\texttt{Irrelevant}]:} {[4]} Lacking Supporting Facts
-{[\texttt{Irrelevant}]:} {[5]} Lacking Supporting Facts
-{[\texttt{Relevant}]:} {[6]} Yi Yi-Yi Yi is a 2000 Taiwanese drama film written and directed by Edward Yang.{\texttt{$<$/eol$>$}} \\
{\bf $<$Generator$>$} (from \texttt{Response Generator: wikimultihop})
\vspace{0.2em}
           \\\hdashline
Yi Yi \\
{[Cite]}: {[2]} {[6]}{\texttt{$<$/eog$>$}}
\\ \midrule
{\bf $<$Instruction$>$} What is the name of this English musician, comedian, and actor, who narrated Dream Street?{\texttt{$<$/eoi$>$}} \\ 
{\texttt{$<$retrieval$>$}}{[1]} Dream Street (UK TV series)
-Dream Street is a British children's television series that ran from 6 May 1999 to 2002 on "CITV". The show is narrated by British comedian Russ Abbot, and was aimed at children aged from 2 to 7. The show featured talking toy vehicles, which were radio controlled in real time. The series aired back on CITV around 2008.
-{[2]} Dream Street (UK TV series) -Dream Street is a British children's television series that ran from 6 May 1999 to 2002 on "CITV". The show is narrated by British comedian Russ Abbot, and was aimed at children aged from 2 to 7. The show featured talking toy vehicles, which were radio controlled in real time. The series aired back on CITV around 2008. -{[3]} Russ Abbot -Russ Abbot (born Russell A. Roberts; 18 September 1947) is an English musician, comedian and actor. He first came to public notice during the 1970s as the singer and drummer with British comedy showband the "Black Abbots", along with Leonard 'Lenny' Reynolds, later forging a prominent solo career as a television comedian with his own weekly show on British television. -{[4]} Dream Street (Janet Jackson album) -nDream Street is the second studio album by American musician Janet Jackson, released on October 23, 1984, by A\&M Records. More pop than her debut album's "bubblegum soul" feel, the album wasn't the runaway success that Janet's father Joseph thought it would be, peaking at number one hundred forty-seven on the "Billboard" 200 in 1984. The album did have one modest hit for Jackson, the top ten R\&B single, "Don't Stand Another Chance", produced by brother Marlon. Also, the video for the song "Dream Street", her first music video, was shot during the shooting of the TV show "Fame".{\texttt{$<$/retrieval$>$}} \\
{\bf $<$Locator$>$}   (from \texttt{Fact Loctor: wikimultihop}) 
\vspace{0.2em}
           \\\hdashline
{[\texttt{Relevant}]:} {[1]} Dream Street (UK TV series) -Dream Street is a British children's television series that ran from 6 May 1999 to 2002 on "CITV".
-{[\texttt{Relevant}]:} {[2]} Dream Street (UK TV series) -The show is narrated by British comedian Russ Abbot, and was aimed at children aged from 2 to 7.
-{[\texttt{Relevant}]:} {[3]} Russ Abbot -Russ Abbot (born Russell A. Roberts; 18 September 1947) is an English musician, comedian and actor.
-{[\texttt{Relevant}]:} {[4]} Lacking Supporting Facts  {\texttt{$<$/eol$>$}}          
\\ 
\bottomrule
 \end{tabular}
    \caption{Training examples of the short-trajectory subset.}\label{tab:examplse_short_training_table}
\end{table*}

\begin{table*}[t!]
\begin{center}
\begin{tcolorbox}
[colback=black!5!white,colframe=gray!15!gray,width=\textwidth,title={Prompt used in ChatGPT for fact locator.}]
\textbf{Instruction:}
Given an \{instruction type\}, answer, and external knowledge passage, your job is to determine whether the passage is relevant to the question and can provide useful information for obtaining the answer. If the passage meets this requirement, respond with [Relevant] and extract useful spans from the passage. The extracted spans consist of complete sentences to make the extracted span understandable standalone.
\tcblower
\textbf{Question}: Who won the 2016 ncaa football national championship?

\textbf{Answer}: The 2015 - 2016 season's ncaa national football championship game was played between the Clemson Tigers and the Alabama Crimson Tide on January 11, 2016. The Alabama Crimson Tide won the game by holding off the undefeated Clemson Tigers 45–40 in the fourth quarter.

\textbf{Passage}: The 2016 College Football Playoff National Championship was a bowl game that determined a national champion of NCAA Division I FBS college football for the 2015 season. It was played at University of Phoenix Stadium in Glendale, Arizona on January 11, 2016, and was the culminating game of the 2015–16 bowl season. The game was played between the winners of two pre-designated semifinal bowls played on December 31, 2015: the No. 1 Clemson Tigers, who beat the No. 4 Oklahoma Sooners 37–17 at the Orange Bowl, coached by Dabo Swinney in his 8th season, and the No. 2 Alabama Crimson Tide, who shut out the No. 3 Michigan State Spartans 38–0 at the 

\textbf{Rating}: [Relevant]

\textbf{Extracted span}: It was played at University of Phoenix Stadium in Glendale, Arizona on January 11, 2016, and was the culminating game of the 2015–16 bowl season. 

\textbf{Question}: Where was three billboards outside ebbing missouri film?

\textbf{Answer}: Principal filming for Three Billboards Outside Ebbing, Missouri was in Sylva, North Carolina. The actual billboards for the film were put in a pasture near Black Mountain, North Carolina, and the billboard scenes were filmed there because that location was better.

\textbf{Passage}: actual sign appeared in the film. The musical score was written by Carter Burwell, who had also supplied the score for McDonagh's films "In Bruges" and "Seven Psychopaths". As well as Burwell's score, the film features songs by ABBA, Joan Baez, The Felice Brothers, the Four Tops, Monsters of Folk, and Townes Van Zandt. "Three Billboards Outside Ebbing, Missouri" premiered in competition at the 74th Venice International Film Festival on September 4, 2017. It also had screenings at the 2017 Toronto International Film Festival, the 2017 San Sebastián International Film Festival (where it won the Audience Award), the BFI London

\textbf{Rating}: [Irrelevant]

\textbf{Extracted span}: None

\textbf{Question}: \{input\}

\textbf{Answer}: \{output\}

\textbf{Passage}: \{text\}

\textbf{Rating}:
\end{tcolorbox}
\end{center}
    \caption{Prompt examples used in ChatGPT for fact locator of long-trajectory subset.}
    \label{tab:examplse_Prompt_fact}
\end{table*}

\begin{table*}[t!]
\begin{center}
\begin{tcolorbox}
[colback=black!5!white,colframe=gray!15!gray,width=\textwidth,title={Prompt used in ChatGPT for WikiMHQA.}]
\textbf{Instruction:}
Given a question, provide knowledge search intent to help better retrieve the answer from external document on the web (e.g., Wikipedia). Split the intent with ';' and write an explanation.
\tcblower
\textbf{Question}: Which magazine was started first Arthur's Magazine or First for Women?

\textbf{Search Intent}: Arthur's Magazine publication year; First for Women publication year

\textbf{Explanation}: By splitting the search into two parts, each focusing on the foundation date of the respective magazines. This approach allows for a direct comparison of their start dates to determine which magazine was started first.

\textbf{Question}: What is the legal age of marriage, without parental consent or other authorization, in Nebraska?

\textbf{Search Intent}:  legal age of marriage in Nebraska without parental consent

\textbf{Explanation}: This intent directly targets the core of the question, ensuring that the search results will specifically address the legal stipulations surrounding marriage age in Nebraska, making it easier to provide a precise answer.

\textbf{Question}: Californian rock band Lit recorded A Place in the Sun in 1995, but what's their best known song?

\textbf{Search Intent}: Californian rock band Lit's most famous and popular songs

\textbf{Explanation}: The search focuses on identifying the most popular or well-recognized song by the Californian rock band Lit. This does not require information about the album A Place in the Sun or its recording date, but rather seeks to find which song by Lit is generally considered their biggest hit or most iconic track.

\textbf{Question}: \{input\}

\textbf{Search Intent}:
\end{tcolorbox}
\end{center}
    \caption{Prompt examples used in ChatGPT for intent reconstructor of long-trajectory subset.}
\label{tab:examplse1_Prompt_intent}
\end{table*}

\begin{table*}[t!]
\begin{center}
\begin{tcolorbox}
[colback=black!5!white,colframe=gray!15!gray,width=\textwidth,title={Prompt used in ChatGPT for Alpaca-GPT4.}]
\textbf{Instruction:}
Given an instruction, provide clarified knowledge search intent to help better retrieve the answer from external document on the web (e.g., Wikipedia). If there are different intents, split them with ';'. 
\tcblower
\textbf{Instruction}: Write a response that appropriately completes the request.\textbackslash n\textbackslash n Instruction:\textbackslash n Name some nations with a monarchy government.

\textbf{Search Intent}: nations with a monarchy government

\textbf{Explanation}: The search focuses on nations with a monarchy government.

\textbf{Instruction}: Tell me two advantages of using AI assistants.?

\textbf{Search Intent}: Advantages of Artificial Intelligence Assistants

\textbf{Explanation}: The question "Tell me two advantages of using AI assistants?" has a search intent focused on understanding the benefits of AI assistants. The query seeks to identify two specific advantages of using AI assistants.

\textbf{Instruction}: Task: Come up with 5 example datasets that demonstrate the use of natural language processing.\textbackslash n $<|$Input$|>$: $<$No input$>$

\textbf{Search Intent}: natural language processing example dataset

\textbf{Explanation}: the instruction asks for 5 example datasets that demonstrate the use of natural language processing (NLP). To determine the search intent, you need to identify the core elements of the question, which are "natural language processing" and "example datasets.

\textbf{Instruction}: \{input\}

\textbf{Search Intent}:
\end{tcolorbox}
\end{center}
    \caption{Prompt examples used in ChatGPT for intent reconstructor of long-trajectory subset.}
\label{tab:examplse2_Prompt_intent}
\end{table*}

\begin{table*}[t!]
\begin{center}
\begin{tcolorbox}
[colback=black!5!white,colframe=gray!15!gray,width=\textwidth,title={Prompt used in ChatGPT for WoW.}]
\textbf{Instruction:}
Given a question, answer and chat history separated by new lines, provide a knowledge search intent for the question to help better obtain answers from external documents on the web (e.g., Wikipedia). The intent needs to consider important and necessary contextual information from history so that it can be fully understood.
\tcblower
\textbf{History}: History: What can you tell me about Gary Cherone?
\textbackslash n Gary Francis Caine Cherone is an American rock singer and songwriter, known for his work as the lead vocalist of Extreme and for his short stint for Van Halen. 
\textbackslash n Did Gary Cherone sing well?
\textbackslash n Yes, Gary Cherone is also known for his work as the lead vocalist of the Boston rock group Extreme.
\textbackslash n What significant fact can you tell me about Gary Cherone that you liked?
\textbackslash n I like that Gary Cherone remained in contact and on good terms with Van Halen.
\textbackslash n What did Gary Cherone do after Van Halen?
\textbackslash n After his departure from Van Halen, Gary Cherone returned to Boston and put together a new project, Tribe of Judah.

\textbf{Question}: Did they release any albums during that time frame?

\textbf{Answer}: After Gary Cherone, Eddie Van Halen recovered from his hip surgery in November 1999, and no official statements were made by Van Halen and no music was released.

\textbf{Explanation}: 
The search intent arises from the user's interest in Gary Cherone's activities after leaving Van Halen. Therefore, the question "Did they release any albums during that time frame?" is interpreted as the user wanting to know if Van Halen released any albums after Cherone's departure.

\textbf{Search Intent}: Any album released by Eddie Van Halen after Gary Cherone left

\textbf{History}: Where does Call of the Dead take place\textbackslash n It takes place in a desolate area of the Siberian tundra next to the frozen ruins of a broken cargo ship and a old Soviet lighthouse.\textbackslash n What is Call ForThe Dead's theme\textbackslash n The players are once again are tasked with surviving the never-ending onslaught of the Zombie hordes, while also dealing with a new, dangerous threat.

\textbf{Question}: What is the genre?

\textbf{Answer}: The genre is crime, spy novel.

\textbf{Search Intent}: the genre of Call For The Dead

\textbf{Explanation}: Based on the history and the content of the question, the search intent is to determine the genre or category to which "Call For The Dead" belongs. The user is seeking to identify the specific classification of this work within entertainment or gaming.

\textbf{History}: \{history\}

\textbf{Question}: \{input\}

\textbf{Answer}: \{output\}

\textbf{Search Intent}:
\end{tcolorbox}
\end{center}
    \caption{Prompt examples used in ChatGPT for intent reconstructor of long-trajectory subset.}
    \label{tab:examplse_Prompt_intent_long}
\end{table*}

\newpage

\end{document}